\documentclass[10pt,twocolumn,letterpaper]{article}

\usepackage{cvpr}              
\usepackage{graphicx}
\usepackage{amsmath}
\usepackage{amssymb}
\usepackage{booktabs}
\usepackage{listings}
\usepackage[accsupp]{axessibility} 
\usepackage{etoc}
\usepackage{multirow}
\usepackage{algorithmic}
\usepackage{algorithm}
\usepackage[pagebackref,breaklinks,colorlinks]{hyperref}
\usepackage[dvipsnames]{xcolor}

\usepackage[capitalize]{cleveref}
\crefname{section}{Sec.}{Secs.}
\Crefname{section}{Section}{Sections}
\Crefname{table}{Table}{Tables}
\crefname{table}{Tab.}{Tabs.}

\newcommand{\methodname}{SLACK\xspace}
\newcommand{\myparagraph}[1]{\vspace{0.1cm}\noindent {\bf #1.}}
\newcommand{\myparnodot}[1]{\vspace{0.1cm}\noindent {\bf #1}}
\newcommand\Dtrain{\mathcal{D}_{\text{train}}}
\newcommand\Dval{\mathcal{D}_{\text{val}}}
\newcommand\Dcal{\mathcal{D}}
\newcommand\Dtest{\mathcal{D}_{\text{test}}}
\newcommand\Ltrain{\mathcal{L}_{\text{train}}}
\newcommand\Lval{\mathcal{L}_{\text{val}}}

\newcommand{\ours}[1]{{\color{black}{#1}}}

\usepackage{pifont}%
\newcommand{\cmark}{\text{\ding{51}}}
\newcommand{\xmark}{\text{\ding{55}}}

\begin{document}

\title{SLACK: Stable Learning of Augmentations \\ with Cold-start and KL regularization}

\author{
    Juliette Marrie$^{1,2}$
    \and
    Michael Arbel$^1$
    \and
    Diane Larlus$^2$
    \and
    Julien Mairal$^1$ \and \\
    $^1$~{Univ.\ Grenoble Alpes, Inria, CNRS, Grenoble INP, LJK} \hspace{1cm} $^2$~{NAVER LABS Europe}
}
\maketitle

\begin{abstract}
Data augmentation is known to improve the generalization capabilities of neural networks, provided that the set of transformations is chosen with care, a selection  often performed manually. 
Automatic data augmentation aims at automating this process.
However, most recent approaches still rely on some prior information; they start from a small pool of manually-selected default transformations that are either used to pretrain the network or forced to be part of the policy learned by the automatic data augmentation algorithm. 
In this paper, we propose to directly learn the augmentation policy without leveraging such prior knowledge. 
The resulting bilevel optimization problem becomes more challenging due to the larger search space and the inherent instability of
bilevel optimization algorithms. To mitigate these issues (i) we follow a successive cold-start strategy with a Kullback-Leibler regularization, and (ii) we parameterize magnitudes as continuous distributions. Our approach leads to competitive results on standard benchmarks despite a more challenging setting, and generalizes beyond natural images.\footnote{Project page: \href{https://europe.naverlabs.com/slack}{https://europe.naverlabs.com/slack}} 
\end{abstract}

\section{Introduction}
\label{sec:intro}

\begin{figure*}
    \centering
    \includegraphics[width=\linewidth]{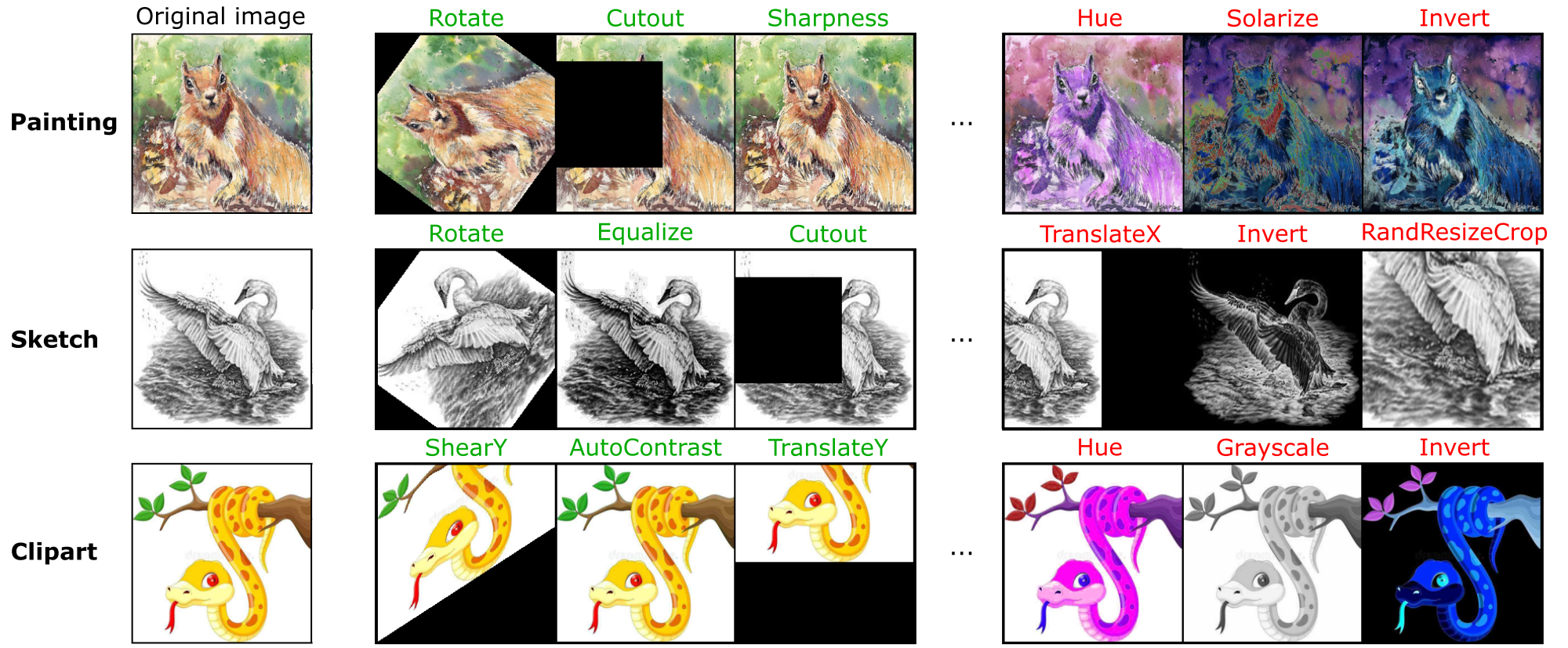}
    \caption{For different domains of the DomainNet dataset~\cite{peng2019domainnet} (one per line), we show an image from that domain (left) and that image transformed using {\color{ForestGreen} the three most likely} (middle) and {\color{red} the three least likely} (right) augmentations for that domain, as estimated by \methodname.}
    \label{fig:eye_catching}
\end{figure*}

Data augmentation, which encourages predictions to be stable with respect to particular image transformations, has become an essential component in visual recognition systems.
While the data augmentation process is conceptually simple, choosing the optimal set of image transformations for a given task or dataset is challenging. For instance, designing a good set for  ImageNet~\cite{deng09imagenet} or even CIFAR-10/100~\cite{krizhevsky2009learning} has been the result of a long-standing research effort.
Whereas data augmentation strategies that have been chosen by hand for ImageNet have been used successfully for many recognition tasks involving natural images, they may fail to generalize to other domains such as medical imaging, remote sensing or hyperspectral imaging.

This has motivated automating the design of data augmentation strategies~\cite{ho19population,hataya22madao,li20dada,lim19fastaa,lin19ohlaa,muller21ta,wang21daas,zheng22deepaa}. Those are often represented as a stochastic \emph{policy} that randomly draws a combination of transformations along with their magnitudes from a large predefined set, each time an image is sampled. {The goal becomes to learn strategies that effectively compose multiple transformations, which is a challenging task given the large search space of augmentations.}

A natural framework for learning the parameters of this policy is that of bilevel optimization. Intuitively, one looks for the best possible policy such that a neural network trained with this policy on a training set (inner problem) generalizes well on a distinct validation set (outer problem). Optimizing the resulting formulation is challenging as the outer problem depends on the solution of the inner problem. 
Classical techniques for solving this bilevel problem, such as unrolled optimization, can become highly unstable as the network weights become progressively suboptimal for the current policy during the learning process.

Moreover, augmentations are often non-differentiable in the parameters of the policy, thus requiring techniques other than direct differentiation, such as Bayesian optimization~\cite{lim19fastaa}, gradient approximations
 (\eg RELAX~\cite{relax}), or the score method / REINFORCE~\cite{reinforce} algorithm. While these techniques bypass the differentiability issues, they can suffer from large bias or variance. 
As a result, learning augmentation policies is a difficult problem whose challenges are exacerbated by the inherent instability of the optimization techniques developed to solve bilevel problems, such as unrolled optimization~\cite{arbel2022non}.

A standard way to improve stability and make the automatic data augmentation problem simpler is to reduce the search space.
This is often achieved by learning the policy on top of ``default'' transformations such as Cutout~\cite{cutout}, random cropping and resizing, or color jittering, all known to be well-suited to natural images which compose standard benchmarks such as CIFAR or ImageNet, or by discarding transformations known to be harmful such as Invert. 
Fixing some of the transformations and removing others mitigate the challenges inherent to learning a composition of transformations. TrivialAugment~\cite{muller21ta} also shows that state-of-the-art results can be achieved on these previous benchmarks simply by directly applying the policy classically used for initializing auto-augmentation models, up to minor modifications. Moreover, all methods rely on carefully chosen ranges that constraint the transformation's magnitudes.
Despite its effectiveness, manually selecting default transformations and magnitude ranges restricts the applicability of such policies to natural images and prevents generalisation to other domains.

In this paper, our goal is to choose augmentation strategies without relying on default transformations nor on hand-selected magnitude ranges known to suit common benchmarks.
To achieve this objective, we first introduce a simple interpretable model for the augmentation policies which allows learning both the frequency by which a given augmentation is selected and the magnitude by which it is applied. Then, we propose a method for learning these augmentation policies by solving a bilevel optimization problem.
Our method relies on the REINFORCE technique for computing the gradient of the policy and on unrolled optimization for learning the policy, both of which can result in instabilities and yield high variance estimates.

To address these issues, we introduce an efficient 
 multi-stage algorithm with a cold-start strategy 
 and a Kullback-Leibler (KL) regularization that are designed to improve the stability of the process for learning the data augmentation policy. 
More precisely, the algorithm first pre-trains a network with a data augmentation policy uniformly sampling over all transformations. Then, each stage uses a ``cold-start'' strategy by restarting from the pre-trained network and performs incremental updates of the current policy.

This multi-stage approach with cold start prevents the network from becoming progressively suboptimal as the policy is updated using unrolled optimization. 
The KL regularization defines a trust region for the policy to compensate for the possibly high variance of
 gradient estimates obtained using the REINFORCE technique and encourages exploration during training, preventing collapse to trivial solutions. 
 This regularization is inspired by proximal point algorithms in convex optimization~\cite{rockafellar1976monotone}, which have also been successful in reinforcement learning tasks~\cite{schulman2017proximal}. 

By combining the regularized multi-stage approach with our interpretable model of the augmentation policies, we obtained the proposed SLACK method, which stands for \emph{Stable Learning of Augmentations with Cold-start and Kullback-Leibler regularization}. 
SLACK is an efficient data augmentation learning method 
that is able to address the challenging bilevel optimization problem of learning a stochastic data augmentation policy without relying strongly on prior knowledge.  
Figure~\ref{fig:eye_catching} illustrates the transformations found by SLACK to be most important / detrimental on a dataset of different domains including non-natural images.

To summarize, our contribution is threefold. 
(i) We propose a simple and interpretable model of the policies which allows learning both frequency and  magnitudes of the augmentations.
(ii) We propose a regularized multi-stage strategy to improve the stability of the bilevel optimization algorithm 
used for solving the data augmentation learning problem.
(iii) We evaluate our method on challenging experimental settings, and show that it finds competitive augmentation strategies on natural images without resorting to prior information and generalizes to other domains.
\section{Related Work}
\label{sec:rw}

The choice of image transformation (also known as data augmentation) has become central in the design of computer vision pipelines. To remove the burden of manual selection,  automatic data augmentation strategies have been proposed~\cite{paulin2014transformation,volpi2019addressing}.
AutoAugment~\cite{cubuk19autoaugment}, one of the earliest methods, uses for instance a recurrent neural network for designing the augmentation policy. Because such an approach requires retraining a prediction model at each iteration, it is prohibitively slow, and more efficient alternatives have been proposed. 
They aim at reducing the training cost using, \emph{e.g.}, population-based training~\cite{ho19population}, Bayesian optimization~\cite{lim19fastaa}, and more recently, gradient-based approaches based on bilevel optimization~\cite{hataya20fasteraa,hataya22madao,li20dada,lim19fastaa,liu21ddas,wang21daas,zhang20advaa}, relying on various gradient estimation techniques such as RELAX~\cite{relax} or the Score method~\cite{reinforce}. While the former is inherently biased, the latter is theoretically exact, but has a high variance when approximated in the context of stochastic optimization. Therefore, these approximations may lead to diverging gradient updates. 
Our method alleviates this by introducing a KL regularization that defines a trust-region for the policy.

\myparagraph{Automatic augmentation using prior knowledge}
Most previous works learn augmentations using a small network learned on a subset of the dataset of interest, before retraining the prediction model on a larger network using the full (augmented) data. 
This choice is appealing to recent gradient-based methods~\cite{hataya20fasteraa, li20dada} as the search phase for an augmentation policy is often reduced to minutes. Nevertheless, \cite{hataya22madao,zhang20advaa} have observed that policies found with such a reduced setup may be suboptimal compared to approaches exploiting full datasets for training both the augmentation policy and the prediction model.
This observation was confirmed in \cite{cubuk20ra}, which shows that a naive grid search could actually yield state-of-the-art results when directly training on the full-size network and the full data. 
These results are however obtained at the expense of using strong prior knowledge: 
augmentation policies are applied on top of default transformations that are manually and independently chosen for each benchmark. 
Lately, \cite{muller21ta} has shown that with a few additional careful choices regarding the augmentation policies, applying a single random transformation on top of the default ones could lead to state-of-the-art results.

To avoid relying on default augmentations, DeepAA~\cite{zheng22deepaa} has recently proposed a greedy approach that is able to learn these transformations. Yet, learning is performed after a ``pre-training'' phase leveraging the usual default transformations. Moreover, while such a greedy approach simplifies the search procedure and reduces its stochasticity, the resulting computational cost is high. Instead, our approach improves stability and allows directly learning the joint probability of sampling multiple transformations, reducing the search time twofold compared to DeepAA.

\section{Method}

\begin{figure}
    \centering
    \includegraphics[width=\linewidth]{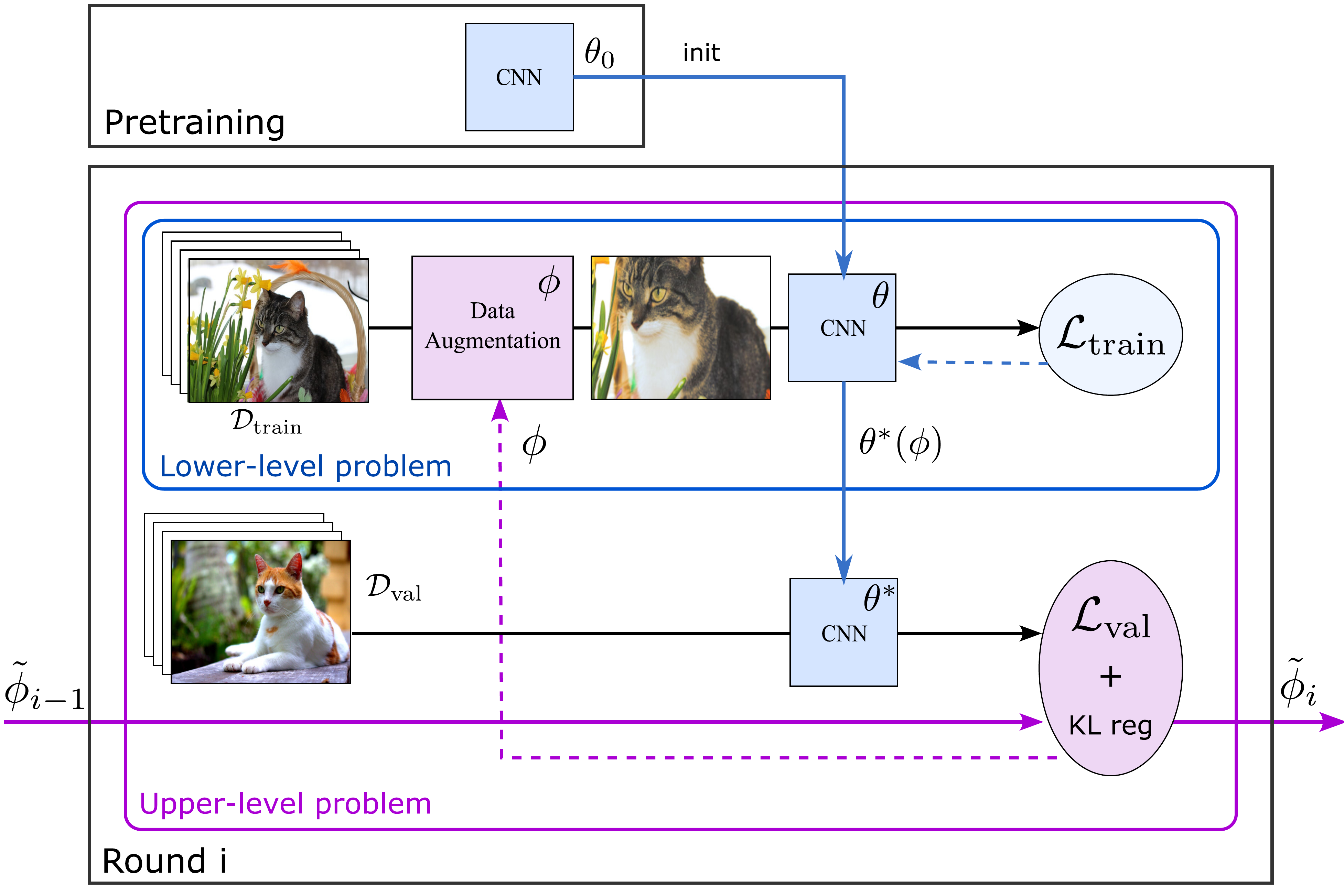}
    \caption{\textbf{Overview of our proposed \methodname~method.}
     We learn a data augmentation policy parameterized by $\phi$ using bilevel optimization. The inner loop finds the optimal network parameter~$\theta^{*}$ on images from $\Dtrain$. The outer loop trains on a disjoint set of images $\Dval$ using this network and finds the optimal transformation parameters $\phi$. The method is enhanced with i) a cold-start strategy that structures the learning into rounds which share~$\phi$ but restart the network from the pretrained one, and ii) a KL regularization.}
    \label{fig:model}
\end{figure}

Our method, \methodname, defines an \emph{augmentation policy}, which is a probabilistic model for generating data augmentations. 
The goal is to learn the parameters of this augmentation policy so as to improve the performance of the trained classifier on a held-out dataset.  
We first describe the augmentation policy (Sec.~\ref{sec:param}) and then formalize the problem of learning data augmentations with bilevel optimization (Sec.~\ref{sec:bilevel}). We then describe our approach for solving such a bilevel optimization problem, which aims at stabilizing the optimization (Sec.~\ref{sec:algorithm}).

\subsection{Stochastic data augmentations policy}
\label{sec:param}

An augmentation function $\tau$ transforms an image $x$ into another \emph{augmented} image $\tau(x)$ of the same dimensions.
We consider composite augmentations obtained by combining simpler augmentations selected from a finite set $\mathcal{S} = \{s_1,\dots,s_N\}$ of $N$ candidate \emph{elementary transformations}, such as rotations, translations, shearing, etc. 
Each elementary transformation depends on a \emph{magnitude} parameter $m$ that controls the strength of the transformations, for instance, the angle by which an image is rotated. Magnitudes are normalized to be in the unit interval $[0,1]$. 

\myparagraph{Augmentation policy} 
We define the augmentation policy as a probabilistic model $p_{\phi}$ that generates composite augmentations given some parameter $\phi$ to be learned. The model generates an augmentation in three steps: 
(1) it samples $K$ elementary transformations $t_1,\dots,t_K$ from $\mathcal{S}$  according to a categorical distribution $p_{\pi}$ of parameter $\pi$, (2) it samples values for the magnitudes $m_1,\dots,m_K$ for each of the selected elementary transformations $t_k$ according to a smoothed uniform distribution $p_{\mu}$ of parameter $\mu$
and (3) it composes the $K$ elementary transformations to obtain the composite augmentation, with each $t_k$ applied using its corresponding magnitude $m_k$. 
Therefore, the augmentation policy $p_{\phi}(\tau)$  takes the form
\begin{align}\label{eq:policy}
	p_{\phi}(\tau) = \prod_{i=1}^K p_{\pi}(t_i) p_{\mu}(m_i|t_i),
\end{align}
where the parameters $\phi=(\pi, \mu)$ are learned jointly.
Next, we describe the sampling of the transformations and of their magnitudes.  
 
\myparagraph{Sampling transformations} 
We sample elementary transformations $t_k$ with replacement from a categorical distribution $\text{Cat}_{\pi_{k}}$ of dimension $N$ parameterized by a logit vector $\pi_k {:=} (\pi_{k,n})_{1\leq n\leq N}$. The probability $p_{\pi}(t_1,\dots,t_K)$ of sampling the $K$ transformations is given by:
  \begin{align}\label{eq:cat}
  	p_{\pi}(t_1,\dots,t_K) = \Pi_{k=1}^K \text{Cat}_{\pi_{k}}(t_k),
  \end{align}
where we collect all logits to form a parameter matrix $\pi$ of size $K\times N$. These parameters are learned.   

\myparagraph{Sampling magnitudes} 
The magnitudes of each elementary transformation $s_i$ in $S$ are sampled from a smoothed uniform distribution between $[0, \mu_i]$ whose upper-bound $\mu_i$ is learned.
More precisely, the distribution's density is defined as
$$
p_{\mu_i}(m_i) = \frac 1\mu_i \int_0^{\mu_i} \mathcal{N}(m_i, \sigma)(u) du,
$$
where $\mathcal{N}(m_i, \sigma)$ is the Gaussian distribution of mean $m_i$ and deviation $\sigma$.  
The density $p_{\mu_i}(m_i)$ approximates the uniform distribution $\frac 1\mu_i \mathbf{1}_{[0,\mu_i]}$ as the deviation $\sigma$ approaches $0$. In practice, we set   $\sigma=0.1$ as we found it to achieve a good trade-off between smoothing and approximation.

\myparnodot{Why a uniform distribution?}
We ran some ablations on previous methods (see supplementary) which suggest that a uniform sampling works on par with more elaborate sampling strategies, and that the magnitude range has more impact on the results.

\subsection{Bilevel formulation for policy search}
\label{sec:bilevel}

We consider a prediction task, such as predicting the class $y$ of some natural image $x$ using a  model $f_{\theta}(x)$ with parameter $\theta$. 
We are interested in finding the best policy parameter $\phi$ so that the prediction model $f_{\theta}$, when trained using such policy on a training set $\Dcal$ of input/output pairs $(x,y)$, generalizes well on the test set $\Dtest$.
The problem naturally decomposes in two phases. During the \emph{search phase}, the optimal augmentation policy $p_\phi$ is learned on~$\Dcal$. During the \emph{evaluation phase}, the model is re-trained on $\Dcal$ using $p_\phi$ and is then  evaluated on $\Dtest$. 
The \emph{evaluation phase} is performed using standard optimization methods. However, the \emph{search phase} requires solving a complex optimization problem that we describe next. 

\myparagraph{Search phase as a bilevel problem} 
The \emph{search phase} naturally writes as a bilevel problem involving two interdependent losses: 
a lower-level loss $\Ltrain(\theta,\phi)$ for learning an optimal model parameter $\theta^{\star}(\phi)$ obtained using the augmentation policy $p_{\phi}$ and an upper-level loss $\mathcal{F}(\phi)$ for learning the policy parameter $\phi$ by evaluating the optimal model with parameter $\theta^{\star}(\phi)$. 
Each of these objectives is evaluated on two separate splits of the available data $\mathcal{D}$:  a training split $\Dtrain$ for the lower-level loss and a validation split $\Dval$ for the upper-level loss. Below, we describe both losses.

\myparagraph{Lower-level loss}
We first introduce the training loss $\ell_{\text{train}}(\theta,\tau)$ when only a fixed augmentation $\tau$ is used:
$$
\ell_{\text{train}}(\theta,\tau):= \mathbb{E}_{(x,y) \sim \Dtrain}\left[ \ell( y, f_\theta( \tau(x))) \right],
$$
where $(x,y)$ is an (image,label) pair drawn from $\Dtrain$ and $\ell$ is a pointwise  prediction loss (\eg cross-entropy). We then define the training loss  $\Ltrain(\theta,\phi)$ for an augmentation policy $p_{\phi}$ by taking the expectation of $\ell_{\text{train}}(\theta,\tau)$ over augmentations $\tau$ sampled according to the policy $p_{\phi}$:
$$
\Ltrain(\theta,\phi) := \mathbb{E}_{ \tau \sim p_\phi}\left[ \ell_{\text{train}}(\theta,\tau) \right].
$$
Hence, for a given policy $p_{\phi}$, the goal is to learn the optimal model parameter $\theta^{\star}(\phi)$ by minimizing $\Ltrain(\theta,\phi)$ over $\theta$. 

\myparagraph{Upper-level loss} 
We first denote by  $\Lval(\theta)$ the validation loss for a given model of parameter $\theta$:
$$
\Lval(\theta) := \mathbb{E}_{(x,y) \sim \Dval}\left[ \ell( y, f_\theta( x)) \right]. 
$$
The validation loss $\Lval(\theta)$ is computed over the validation set $\Dval$ \emph{without} applying any augmentation and thus provides a proxy for the performance on the test dataset. 
We then define the upper-level loss to be the validation loss of an optimal model $\theta^{\star}(\phi)$ learned using a policy $p_{\phi}$:
\begin{equation}\label{eq:upper-level}
\mathcal{F}(\phi) := \Lval(\theta^{\star}(\phi)).
\end{equation}

While optimizing the lower-level loss is relatively standard, minimizing the upper-level loss $\mathcal{F}$ is more challenging due to the complex dependence of the optimal model parameter $\theta^{\star}(\phi)$ on the policy. Next, we describe our proposed algorithm for solving the bilevel problem. 

\subsection{SLACK algorithm}\label{sec:algorithm}

We propose \cref{alg:slack} for learning the optimal policy during the search phase. 
SLACK first pre-trains the prediction model using the objective $\Ltrain(\theta,\phi_{\text{uniform}})$ for an initial policy parametrized by $\phi_{\text{uniform}}$ which samples uniformly among all elementary transformations. It then  performs $n_{\text{rounds}}$ rounds to update the parameters $\theta$ and $\phi$ jointly using a bilevel optimization algorithm.  
This approach is reminiscent of the one of AutoAugment (AA)~\cite{cubuk19autoaugment} that fully re-trains the network for each policy update. Yet, it is several orders of magnitude faster than AA, as it benefits from pre-training and from our bilevel optimization.

SLACK relies on two strategies to ensure the stability of the parameter updates during each round: a \emph{cold-start strategy}  for the prediction model and an \emph{anchoring strategy} for the policy. 
The \emph{cold-start}  initializes the prediction model at the beginning of each round using a pre-trained model $\theta_0$. 
\emph{Anchoring} is achieved by encouraging the current policy to remain close to some \emph{anchor policy} $p_{\tilde{\phi}}$. We set $\tilde{\phi}$ to the current policy parameter $\phi$ at the beginning of each round.
During the first $n_{\text{retrain}}$ steps of each round, the algorithm only updates the model parameter using a stochastic estimate $\hat{g}_{\theta}$ of $\nabla_{\theta}\Ltrain(\theta,\phi)$ while maintaining the policy fixed. Then for the last $n_{\text{total}} - n_{\text{retrain}}$ steps, the algorithm alternates between model updates and policy updates. The policy updates aim to minimize the sum of the upper-level objective $\mathcal{F}$ and an anchoring $d(p_\phi, p_{\tilde\phi}) := \text{KL}(p_\pi, p_{\tilde{\pi}})$ encouraging the policy $p_{\phi}$ to remain close to the \emph{anchor policy} $p_{\tilde{\phi}}$. These updates are obtained using a stochastic estimate $\hat{G}_{\phi}$ along with the exact gradient of the KL regularization which admits a closed-form expression. Next we explain how we estimate the gradients $\hat{g}_{\theta}$ and $\hat{G}_{\phi}$ and discuss the effect of cold-start and KL-regularization.

\begin{algorithm}[H]
\footnotesize
\caption{SLACK}\label{alg:slack}
	\begin{algorithmic}[1]
		\STATE {\bf Initialize policy parameter $\phi\leftarrow\phi_{\text{uniform}}$.} 
		\STATE {\bf Pre-training:} $\theta_0\leftarrow$ \verb+optimize+ $\left(\Ltrain(\theta,\phi)\right)$.
		\FOR{ $i \in \{1,...,n_{\text{rounds}}\}$}
                        	\STATE {\bf Cold-start:} $\theta \leftarrow \theta_0$.
                        \STATE {\bf Update anchor policy:} $\tilde{\phi}\leftarrow \phi$.
			\FOR{ $j \in \{1,...,n_{\text{total}}\}$}
	\STATE { Compute stochastic gradient $\hat{g}_{\theta} \approx \nabla_{\theta}\Ltrain(\theta,\phi)$}	
    \STATE { Update $\theta$:} $\theta\leftarrow \theta-\eta \hat{g}_\theta$.
			\IF{ $j>n_{\text{retrain}}$}
    \STATE {Compute stochastic gradient $\hat{G}_{\phi} \approx \nabla_{\phi}\mathcal{F}(\phi)$}			
    \STATE  {Update $\phi$:} 
				$\phi\leftarrow \phi-\alpha(\hat{G}_{\phi}+ \lambda\nabla_{\phi}d(p_{\phi},p_{\tilde{\phi}}))$.
			\ENDIF
			\ENDFOR
		\ENDFOR
	\end{algorithmic}
\end{algorithm}

\myparagraph{Gradient estimation}
\cref{alg:slack} requires estimating the gradient of $\mathcal{F}(\phi)$, which is challenging given the complex dependence of the upper-level loss on the policy $p_{\phi}$ through the optimal model parameter $\theta^{\star}(\phi)$ learned using such a policy. In line with previous works~\cite{hataya20fasteraa,li20dada,liu21ddas,wang21daas}, we approximate the optimal model parameter $\theta^*(\phi)$ with a simpler function $\hat{\theta}(\phi)$  that is easier to compute:
\begin{align}\label{eq:approxtheta}
\hat{\theta}(\phi)  :&= \theta - \eta \nabla_\theta \Ltrain(\theta, \phi).
\end{align}
\cref{eq:approxtheta} corresponds to one gradient step to optimize the lower-level loss starting from the current parameter $\theta$ and $\phi$ and using step-size $\eta>0$.  
By keeping track of the dependence in $\phi$ and exploiting the fact that the augmentation policy $p_{\phi}$ has a score $\nabla_{\phi} \log p_{\phi}(\tau)$ that can be computed explicitly using \cref{eq:policy}, we can use the REINFORCE/Score method \cite{fu2006gradient} to derive a closed-form expression for $\nabla_\phi \hat{\theta}(\phi)$ which will serve for approximating the gradient of $\mathcal{F}$: 
\begin{align*}
\nabla_\phi \hat{\theta}(\phi) = -\eta
\mathbb{E}_{\tau \sim p_\phi} \left[\nabla_{\theta}\ell_{\text{train}}(\theta,\tau)\nabla_\phi \log p_\phi(\tau)^{\top} \right].
\label{exp-reinforce}
\end{align*}
Then, we approximate the upper-level loss $\mathcal{F}(\phi)$ with a simpler function  $\hat{\mathcal{F}}(\phi) {:=} \Lval(\hat{\theta}(\phi))$ and the gradient $\nabla_{\phi} \mathcal{F}(\phi)$ with $\nabla_{\phi}\hat{\mathcal{F}}(\phi)$ which is obtained using the chain rule:
\begin{equation}\label{eq:gradient_outer} \nabla_{\phi} \mathcal{F}(\phi) \approx  \nabla_{\phi} \hat{\mathcal{F}}(\phi) =  \nabla_{\theta} \Lval(\hat{\theta}(\phi))^{\top} \nabla_\phi\hat{\theta}(\phi).
\end{equation}
The above expression requires only first-order derivatives and matrix-vector products, which is amenable to efficient implementation using automatic differentiation softwares.

\myparagraph{Stochastic gradient estimates} In practice, we replace all expectations by  estimates on a batch of data and sampled augmentations. More precisely, to compute the approximation $\hat{g}_{\theta}$ to $\nabla_{\theta}\Ltrain(\theta,\phi)$, 
we sample $B_{\text{aug}}$ augmentations from $p_{\phi}$ and then apply each of them to a batch of training data $B_{\text{train}}$ from $\Dtrain$. Using the same batch of data and augmentation, we approximate $\hat{\theta}(\phi)$ and $\nabla_{\phi}\hat{\theta}(\phi)$ appearing in \cref{eq:gradient_outer}. 
Finally, we use a batch $B_{\text{val}}$ of data from $\Dval$ to estimate $\nabla_\theta \Lval(\hat\theta(\phi))$ and compute $\hat{G}_{\phi}$, which is a stochastic estimate of  $\nabla_{\phi}\hat{\mathcal{F}}(\phi)$ in \cref{eq:gradient_outer}.

\myparagraph{Cold-start}
The cold-start strategy allows to re-train the model at each round with the current augmentation policy starting from the pre-trained model.
This approach is closer to the original bilevel formulation which implies finding an optimal prediction model for each policy. 
Initializing with a pre-trained model yields computational gain as fewer iterations are needed to optimize the model.
We could instead use a \emph{warm-start} strategy which initializes the model at each round with the learned model at the previous round. Yet we experimentally observe that such approach progressively leads to overfitting and degrades the quality of the learned policies (see supplementary).

\looseness=-1
\myparagraph{Anchoring using KL regularization}
We experimentally found that adding an anchoring $d(p_\phi, p_{\tilde{\phi}}) := \text{KL}(p_{\pi},p_{\tilde{\pi}})$ with strength parameter $\lambda$ when updating the policy 
prevents the algorithm from collapsing towards trivial policies. 
The anchoring affects only the categorical distribution $p_{\pi}$. For the magnitudes $p_{\mu}$, we did not use anchoring as it is ill-defined for a uniform distribution. Instead, we simply used smaller step-sizes.
Our approach takes inspiration from Proximal Policy Optimization \cite{schulman2017proximal} used in the context of reinforcement learning which is known to improve policy search.

\section{Experiments}
\label{sec:exp}

In this section, we first briefly describe our experimental setup (Sec~\ref{sec:exp_setup}). Then we evaluate our approach on several standard benchmarks composed of natural images (Sec~\ref{sec:sota}) as well as on a benchmark with other domains (Sec~\ref{sec:exp_domainnet}). We finally report some ablation studies (Sec~\ref{sec:exp_ablation}).

\subsection{Experimental setup}
\label{sec:exp_setup}

\myparagraph{Benchmarks}
We first evaluate our model on three standard benchmarks, CIFAR10~\cite{krizhevsky2009learning}, CIFAR100~\cite{krizhevsky2009learning} and ImageNet-100~\cite{tian2019contrastive}, all composed of natural images. To study how well our method generalizes beyond natural images, we also evaluate on the DomainNet dataset~\cite{peng2019domainnet}, which contains 345 classes for 6 different domains. 
To ensure our protocol uses a similar number of training images for each domain, we use a reduced set of 50,000 training images for the two largest domains (real, quickdraw) and leave the remaining images for testing. For the other domains, we isolate 20\% of the data for testing.

\myparagraph{Architectures}
CIFAR10/100 are evaluated with two architectures that are standard for automatic data augmentation: WideResNet-40x2 and WideResNet-28x10~\cite{zagoruyko2016wide}. Unlike previous works whose search phase is only conduced with the smaller WideResNet-40x2, we search and evaluate with the same architecture, as we found it to be better (see Sec. \ref{sec:exp_ablation}).
ImageNet-100 and DomainNet are evaluated with a ResNet-18~\cite{he16resnet} architecture.

\myparagraph{Transformation space}
Our data augmentation search space is composed of the standard pool of 15 transformations:
\textit{Identity, ShearX, ShearY, TranslateX, TranslateY, Rotate, AutoContrast, Equalize, Invert, Solarize, Posterize, Contrast, Brightness, Sharpness, Color}. We add to this pool the transformations that previous methods usually apply by default: Cutout and RandomCrop for CIFAR, RandomResizeCrop for ImageNet, Grayscale for DomainNet. Following standard practice, when RandomResizeCrop is sampled, it is always applied first. We learn the range of its scale parameter. We do not add ColorJitter that is also applied by default in prior work for ImageNet, as it is already a mix of Brightness, Contrast and Color. However we add Hue, which is one component of ColorJitter and never applied by default.
Following prior work, the magnitudes are mapped to [0,1]. After mapping, $\mu$ is initialized at $0.75$ to favour exploration (see details in supplementary).
We also uniformly sample magnitudes for Cutout and RandomCrop, whereas their value is hand-picked in prior work.
Since the datasets are horizontally symmetric, we follow common practice and apply flip by default.

\myparagraph{Policy search}
We apply a train/val split of 0.5/0.5, meaning that half of the data is used to train the model parameters while the other half is used to learn the augmentation policy. 
Pre-training is done in the same setting as the evaluation (see next paragraph), except that we train only with the train data in the train/val split of the search phase.
We use SGD with momentum for the optimization of the validation and training losses. For the latter, we use the same weight decay as for the final policy evaluation.
We sample 8 different augmentations for computing the expectation that is needed for the stochastic gradient estimate, as detailed in Sec.~\ref{sec:algorithm}.

\myparagraph{Policy evaluation}
We evaluate our models following the framework of TrivialAugment~\cite{muller21ta}.
The corresponding hyperparameters can be found in the supplementary. We evaluate each policy with 4 independent runs, meaning that our results are averaged over a total of $4\times 4=16$ evaluations. Our Uniform policy (corresponding to SLACK's initialization) and our reported results on TrivialAugment are evaluated with 8 independent runs. We also report a confidence interval which contains the true mean with probability
$p=95\%$, under the assumption of normally distributed accuracies.

\subsection{Comparison with the state of the art}
\label{sec:sota}

We compare our method with a Uniform augmentation policy as well as many previous approaches for data augmentation, including AutoAugment (AA)~\cite{cubuk19autoaugment}, Fast AutoAugment (FastAA)~\cite{lim19fastaa}, Differentiable Automatic Data Augmentation (DADA)~\cite{li20dada}, RandAugment (RA)~\cite{cubuk20ra}, Teach Augment~\cite{suzuki2022teachaugment}, UniformAugment~\cite{paulin2014transformation}, TrivialAugment (TA)~\cite{muller21ta}, and Deep AutoAugment (DeepAA)~\cite{zheng22deepaa}. 

For each method, we indicate the total number of composed transformations, and the number of hard-coded transformations among those (Tables~\ref{tab:sota} and \ref{tab:imagenet-100}). For SLACK, we evaluate the policies obtained from 4 independent search runs (each with 4 different train/val splits) to assess the robustness of our approach. We follow the same process when reproducing DeepAA on CIFAR10/100.  
Note that all previous methods use a single run for search, before evaluating the policy with one or multiple runs. We report 95\% confidence intervals for those evaluating with multiple runs.

The supplementary provides qualitative results showing the evolution of the probability distributions over the transformations and the final estimated policies for all datasets.

\begin{table*}[t!]
    \footnotesize
    \centering
    \footnotesize
    \begin{tabular}{lccllllll}
    \toprule
    & \multicolumn{2}{c}{\# Augmentations} & \multicolumn{2}{c}{CIFAR10} & \multicolumn{2}{c}{CIFAR100} \\
    \cmidrule(lr){2-3} \cmidrule(lr){4-5} \cmidrule(lr){6-7}
    & Total & Hard-coded &  WRN-40-2 & WRN-28-10 & WRN-40-2 & WRN-28-10  \\
    \toprule
    AA~\cite{cubuk19autoaugment} & 4 & 2 & 96.3 & 97.4 & 79.3 & 82.9 \\ 
    FastAA~\cite{lim19fastaa} & 4 & 2 & 96.4 & 97.3 & 79.4 & 82.7 \\
    DADA~\cite{li20dada} & 4 & 2 & 96.4 & 97.3 & 79.1 & 82.5 \\
    RA~\cite{cubuk20ra} & 4 & 2  & - & 97.3 & - & 83.3 \\
    TeachA ~\cite{suzuki2022teachaugment} & 4 & 2 & - & 97.5 & - & 83.2 \\
    UniformAugment~\cite{lingchen2020uniformaugment}  & 4 & 2 & 96.25 & 97.33 & 79.01 & 82.82 \\
    TA (Wide)~\cite{muller21ta} & 3 & 2 & 96.32 $\pm$ .05 & 97.46 $\pm$ .06 & 79.86 $\pm$ .19 & 84.33 $\pm$ .17  \\
    \midrule 
    \ours{Uniform policy} & 3 & 0 &  96.12 $\pm$ .08 & 97.26 $\pm$ .07 & 78.79 $\pm$ .25 & 82.82 $\pm$ .24 \\
     DeepAA~\cite{zheng22deepaa} & {\color{white}{**}}$6^{**}$ & {\color{white}{*}}$0*$ & - & 97.56 $\pm$ .14 & - & 84.02 $\pm$ .18 \\
   \ours{DeepAA (reproduced)}$^{\dagger}$ & {\color{white}{**}}$6^{**}$ & {\color{white}{*}}$0*$ & 96.25 $\pm$ .11 & 97.27 $\pm$ .11 & 79.26 $\pm$ .35 & 83.38 $\pm$ .33 \\
    \ours{\bf{\methodname (Ours)}} & 3 & 0  & 96.29 $\pm$ .08 & 97.46 $\pm$ .06 & 79.87 $\pm$ .11 & 84.08 $\pm$ .16 \\
    \bottomrule
    \end{tabular}
    \caption{Test accuracies on CIFAR10 and CIFAR100. For \methodname and DeepAA (reproduced) we conduct 4 independent searches, and evaluate each policy with 4 evaluation runs, {meaning that we report averages over 16 evaluations}.  
    TA and DeepAA are also evaluated with multiple evaluation runs. 
    Results for the remaining methods are reported from the corresponding papers and based on a single run.}
    \vspace{-0.1cm}
    \begin{flushleft}
    *: DeepAA uses hard-coded transformations for pre-training.
   **: DeepAA learns random flipping unlike other baselines.
   $\dagger$ We evaluate the policies found from 4 independent search runs as we do for \methodname, using the code from the authors and following their recommendations.
    \end{flushleft}
    \label{tab:sota}
\end{table*}

\myparagraph{CIFAR}
In Table~\ref{tab:sota}, we observe that, despite not hard-coding Cutout and RandomCrop in our policy, our method is competitive on both CIFAR10 and CIFAR100. 

We found that, in general, Cutout and Rotate are selected with a high probability, while the Invert transformation is systematically discarded (see supplementary). This is consistent with 
the choices made in practice by prior work of adding/removing these transformations manually.

We observe a mismatch between DeepAA's reported results~\cite{zheng22deepaa}, and those we obtain when evaluating their approach on multiple search runs, using the author's code and following their recommendations. This is likely due to the stochasticity of the search procedure.

\myparagraph{ImageNet-100}
Results for ImageNet-100 are reported in Table \ref{tab:imagenet-100}. We compare SLACK to our Uniform policy and to TrivialAugment (RA) and (Wide) variants, the latter using larger magnitude ranges for its random transformation.
SLACK's results lie in between both variants and improve over our Uniform policy.

Interestingly, for ImageNet-100, we found that RandomResizeCrop is not favoured during the search phase (see supplementary), suggesting that it is not critical for ImageNet-100. Instead the performance gap between TA (Wide) and TA (RA) suggest that harder transformations are key to a better performance for this dataset.

\begin{table}[t!]
  \centering
  \footnotesize
  \begin{tabular}{lccc}
    \toprule
     & \multicolumn{2}{c}{\# Augmentations} & ImageNet-100 \\
    \cmidrule(lr){2-4} 
    & Total & Hard-coded & ResNet18 \\
    \midrule
    \ours{TA (RA)}~\cite{muller21ta}$^{\dagger}$ & 5 & 4 & 85.87 $\pm$ .30 \\
    \ours{TA (Wide)}~\cite{muller21ta}$^{\dagger}$ & 5 & 4 & 86.39 $\pm$ .18 \\
    \midrule
    \ours{Uniform policy} & 3 & 0 & 85.78 $\pm$ .32 \\
    \ours{\bf{\methodname}} & 3 & 0 & 86.06 $\pm$ .11 \\   
    \bottomrule
\end{tabular}
\caption{Test accuracies on ImageNet-100.}
\label{tab:imagenet-100}
\end{table}

\subsection{Beyond natural images}
\label{sec:exp_domainnet}

For the DomainNet dataset, we compare SLACK to a Uniform policy, to the augmentations used by DomainBed \cite{gulrajani2020domainbed} for domain generalization, and to the TrivialAugment (RA) and (Wide) methods with their ImageNet and CIFAR default settings. Results can be found in Table \ref{tab:domainnet}. 

DomainBed uses the same default transformations as TA ImageNet together with Grayscale, but with smaller magnitudes and unlike TA, does not add a random transformation.
Yet it strongly overfits and performs much lower than TA.
This suggests that augmentations well suited for domain generalization do not perform well on the individual tasks. 
TA (Wide) ImageNet consistently outperforms all other TA flavors. This
further justifies the need to learn the magnitude range and to
eliminate any manual range selection process. 

\begin{figure*}[t!]
    \centering
    \begin{subfigure}{0.32\linewidth}                       \includegraphics[width=\linewidth]{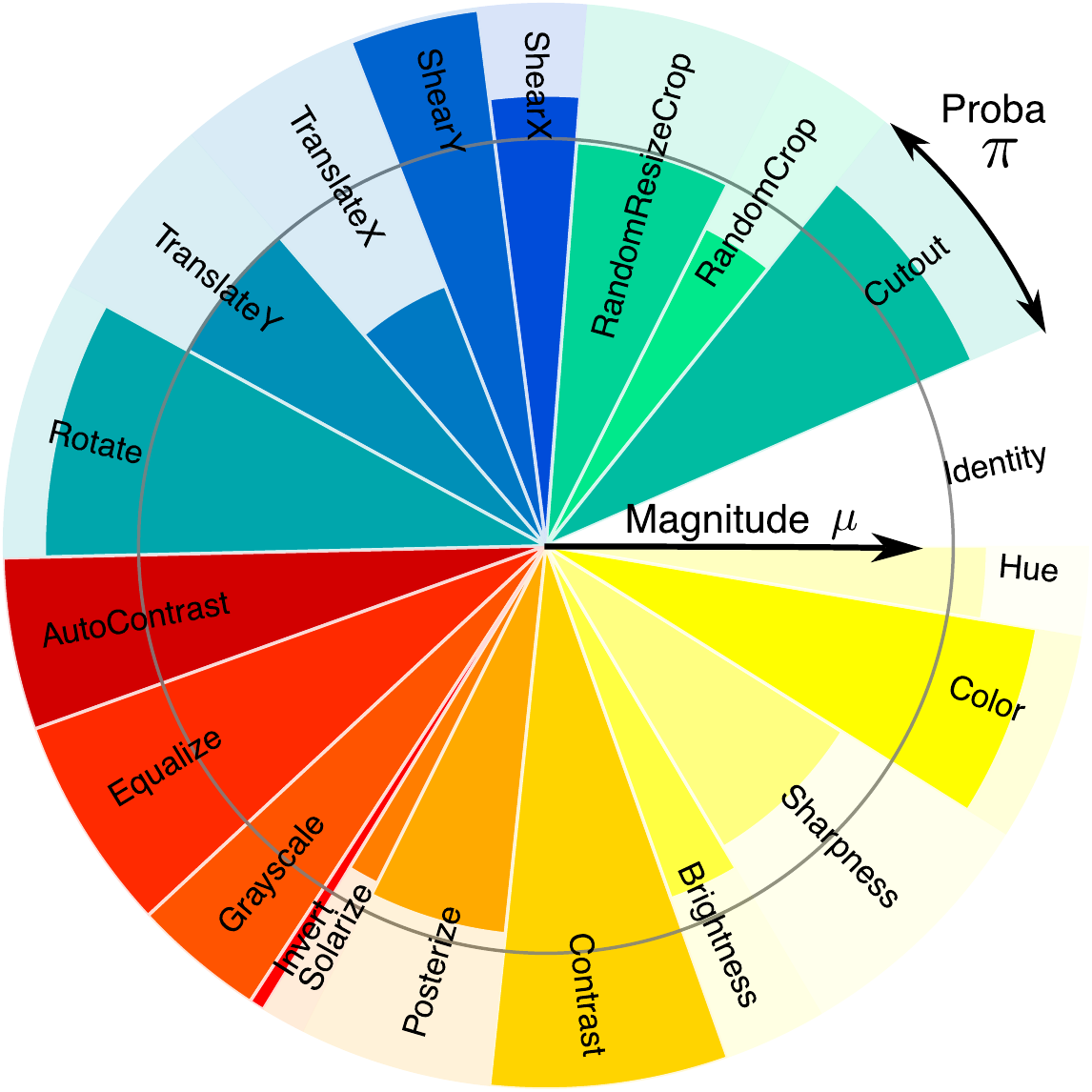}
        \subcaption{Painting}      
    \end{subfigure}
    \begin{subfigure}{0.32\linewidth}                \includegraphics[width=\linewidth]{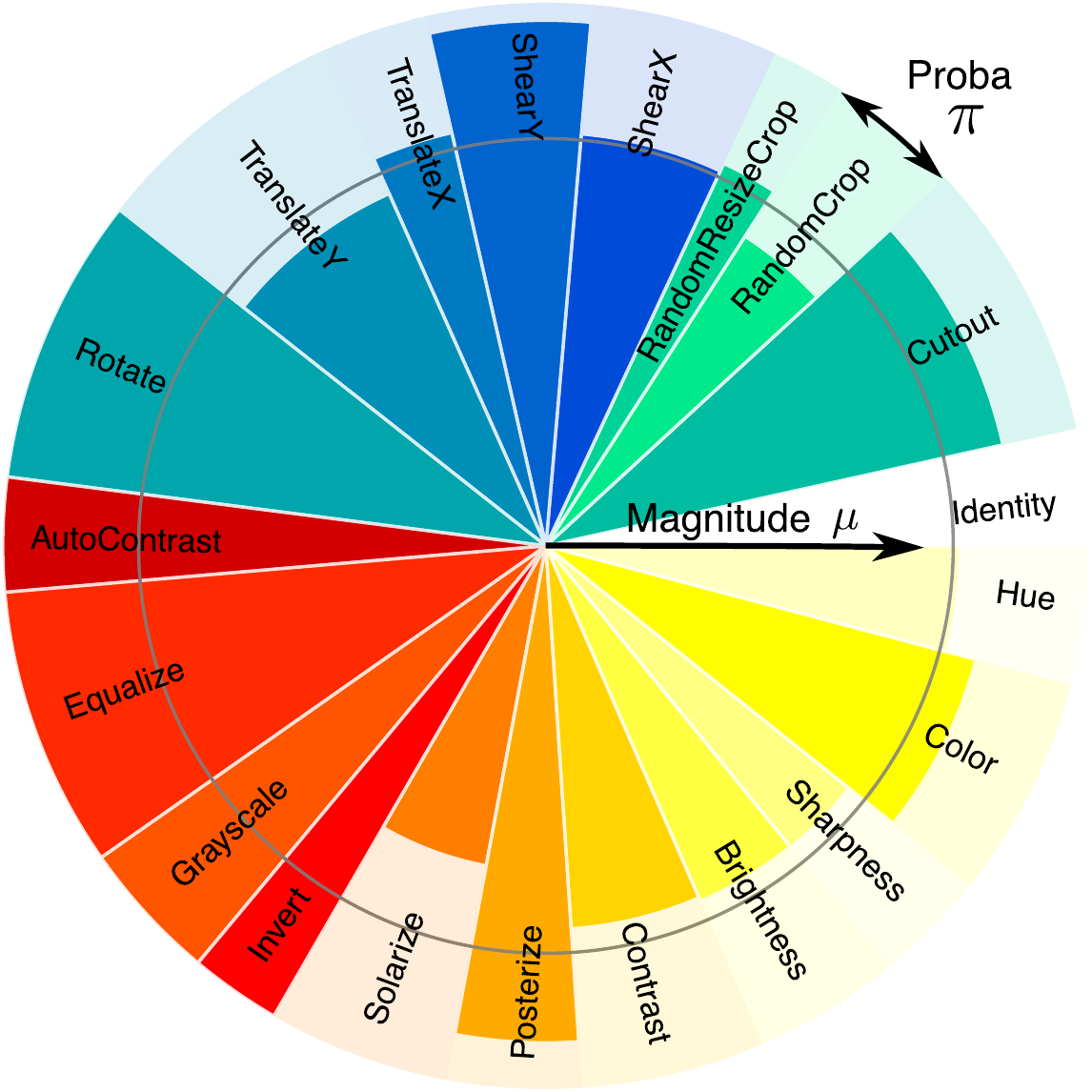}
        \subcaption{Sketch}      
    \end{subfigure}
    \begin{subfigure}{0.32\linewidth}                       \includegraphics[width=\linewidth]{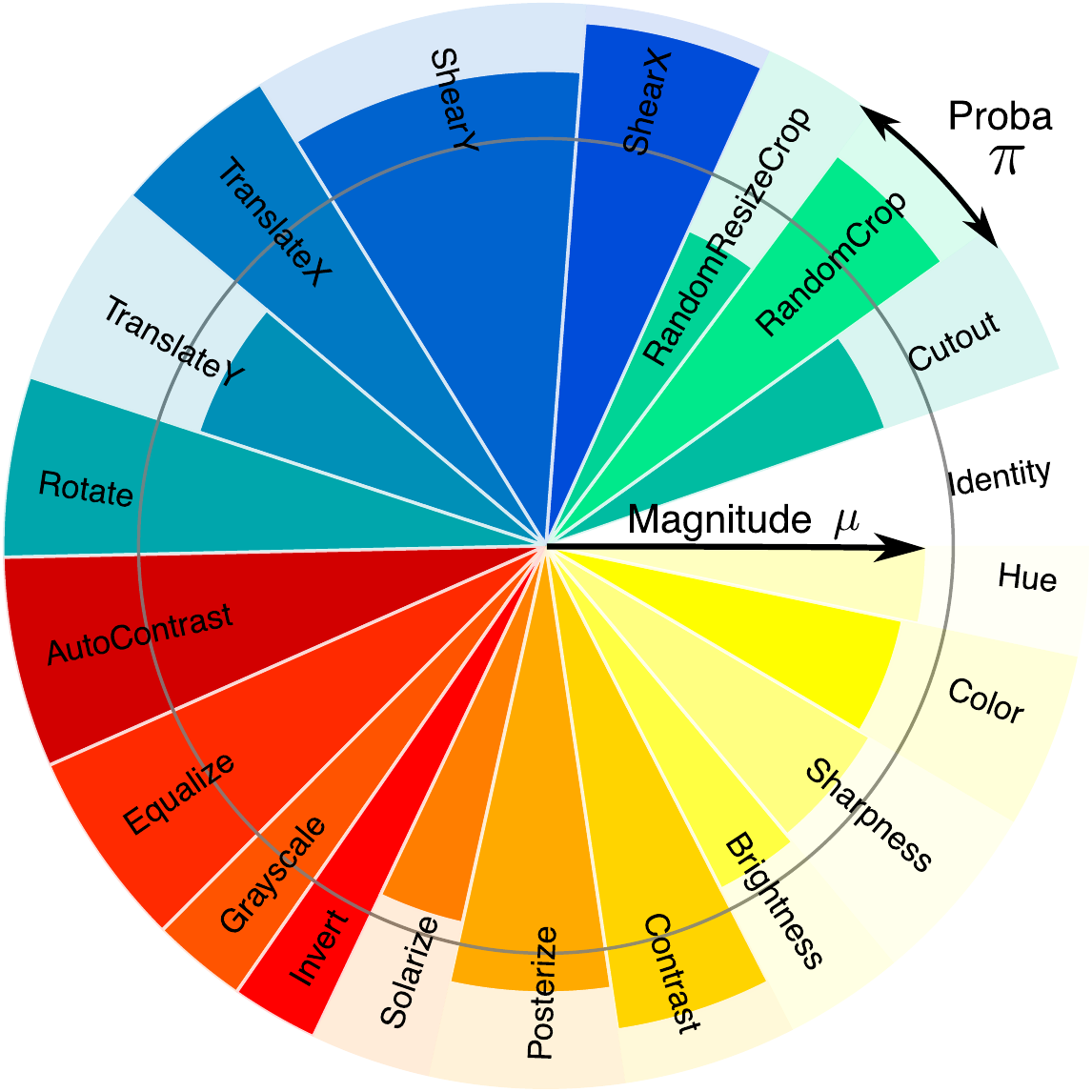}
        \subcaption{Clipart}      
    \end{subfigure}
    \vspace{-0.2cm}
    \caption{Policies found on DomainNet for the best search split. Gray circle: initial magnitude upper-bounds. Radius of each pie: learned upper-bounds. Size of each pie: probability of each transformation, averaged over the three composite distributions. Transformations which are parameter-free, \textit{AutoContrast, Equalize, Grayscale, and Invert,} are displayed with maximal magnitude upper-bound.}
    \label{fig:domainnet-pie}
\end{figure*}

SLACK is a close second, yet it learns the policy end-to-end.
The learned policies are illustrated as pie charts in Fig.~\ref{fig:domainnet-pie}. The slices represent the probability $\pi$ over the different transformations while their radius represent the corresponding magnitudes. They differ from a domain to another, suggesting that the gain compared to the initialization (\ie Uniform policy) results from SLACK's ability to learn and adapt to each domain.
\begin{table*}[h!]
  \centering
  \scalebox{0.97}{
  \scriptsize
  \begin{tabular}{lccccccccc}
    \toprule
    & \multicolumn{2}{c}{\# Augmentations}  & Real-50k & Quickdraw-50k & Inforgraph & Sketch & Painting & Clipart & Average  \\ 
     \cmidrule(lr){2-3} \cmidrule(lr){4-10}
     & Total & Hard-coded &  &  &  \\
    \midrule
    \ours{DomainBed}$^{\dagger}$ & 5 & 5 & 62.54 $\pm$ .15 & 66.54 $\pm$ .91 & 26.76 $\pm$ .36  &  59.54 $\pm$ .37 & 58.31 $\pm$ .25 & 66.23 $\pm$ .10 & 57.23 $\pm$ .18  \\   
    \ours{TA (RA) ImageNet}$^{\dagger}$ & 5 & 4 & 70.85 $\pm$ .13  & 67.85 $\pm$ .07 & \textbf{35.24 $\pm$ .19}  & 65.63 $\pm$ .11  & 64.75 $\pm$ .18  & 70.29 $\pm$ .18 & 62.43 $\pm$ .05 \\
    \ours{TA (Wide) ImageNet}$^{\dagger}$ & 5 & 4 & \textbf{71.56 $\pm$ .07} & 68.60 $\pm$ .05 & \textbf{35.44 $\pm$ .33} & \textbf{66.21 $\pm$ .16} & \textbf{65.15 $\pm$ .20} & 71.19 $\pm$ .19 & \textbf{63.03 $\pm$ .07}  \\
    
    \ours{TA (RA) CIFAR}$^{\dagger}$ & 3 & 2 & 70.28 $\pm$ .08  & 68.35 $\pm$ .07 & 33.85 $\pm$ .21 & 64.13 $\pm$ .12 & 64.73 $\pm$ .17 & 70.33 $\pm$ .21 & 61.94 $\pm$ .05 \\ 
    \ours{TA (Wide) CIFAR}$^{\dagger}$ & 3 & 2 & 71.12 $\pm$ .10  & \textbf{69.29 $\pm$ .05}  & 34.21 $\pm$ .29 & 65.52 $\pm$ .25 & 64.81 $\pm$ .14 & 71.01 $\pm$ .21 & 62.66 $\pm$ .07 \\ 
    \midrule
    \ours{Uniform policy} & 3 & 0 & 70.37 $\pm$ .08 & 68.27 $\pm$ .06 & 34.11 $\pm$ .21 & 65.22 $\pm$ .17 & 63.97 $\pm$ .24 & 72.26 $\pm$ .14 & 62.37 $\pm$ .06 \\
    \ours{\methodname (ours)} & 3 & 0 & 71.00 $\pm$ .13 & 68.14 $\pm$ .11 & 34.78 $\pm$ .18 & 65.41 $\pm$ .16 & 64.83 $\pm$ .12 & \textbf{72.65 $\pm$ .20} & 62.80 $\pm$ .06 \\
    \bottomrule
\end{tabular}
}
\vspace{-0.1cm}
\caption{Test accuracies on DomainNet. 
}
\label{tab:domainnet}
\end{table*}

\subsection{Ablation study}
\label{sec:exp_ablation}

In this section, we evaluate our contributions and main design choices: the network architecture used for search, the KL regularization, and the benefits of learning $\pi$ and $\mu$. More ablations can be found in the supplementary. Note that hyperparameters are adjusted to each baseline included in the comparison, to make them as competitive as possible. 

\myparagraph{Network architecture for search}
In prior works, the search phase (when there is one) is conducted on the smaller WideResNet-40x2 architecture for CIFAR10 and CIFAR100, and the learned policy is evaluated for both WideResNet-40x2 and WideResNet-28x10. Table \ref{tab:search-architecture} shows that for SLACK, searching directly with WideResNet-28x10 gives the best results for that architecture.

\begin{table}[t!]
  \centering
  \scriptsize
  \begin{tabular}{lcc}
     \toprule
     Search architecture & CIFAR10 & CIFAR100 \\
     \midrule
    WRN-40-2 & 97.43 $\pm$ .04 & 83.94 $\pm$ .20  \\
    WRN-28-10 (ours) & 97.46 $\pm$ .06 & 84.08 $\pm$ .16 \\
    \bottomrule
\end{tabular}
\caption{CIFAR10/100 accuracy evaluated with WRN-28-10: impact of using a smaller architecture for the search phase.}
\label{tab:search-architecture}
\end{table}
 
\myparagraph{KL regularization}
We compare \methodname with a flavor that does not apply KL-regularization.
For the latter, we reduce the outer learning rate so that the augmentation policies with and without regularization evolve at similar speeds. 
Results in Table~\ref{tab:ablation-KL} show {that} our regularization is beneficial.

\begin{table}[t!]
  \centering
  \scriptsize
  \begin{tabular}{lcccc}
     \toprule
     & \multicolumn{2}{c}{CIFAR10} & \multicolumn{2}{c}{CIFAR100} \\
    \cmidrule(lr){2-3} \cmidrule(lr){4-5}
    \methodname variant & WRN-40-2 & WRN-28-10 & WRN-40-2 & WRN-28-10 \\
     \midrule
without KL & 96.27 $\pm$ .05 & 97.06 $\pm$ .11 & 79.61 $\pm$ .13 & 83.79 $\pm$ .19  \\
with KL (ours) & 96.29 $\pm$ .08 & 97.46 $\pm$ .06 & 79.87 $\pm$ .11 & 84.08 $\pm$ .16 \\
    \bottomrule
\end{tabular}
\caption{CIFAR10/100 accuracy with/without KL regularization.}
\label{tab:ablation-KL}
\end{table}

\myparagraph{Joint learning of $\pi$ and $\mu$}
Lastly, we study how beneficial jointly learning our augmentation parameters is compared to the initial Uniform policy and to a setting where only $\pi$ or $\mu$ is learned. Results can be found in Table~\ref{tab:norm}.

\begin{table}[t!]
  \centering
  \scalebox{0.97}{
  \scriptsize
  \begin{tabular}{lcccc}
     \toprule
     & \multicolumn{2}{c}{CIFAR10} & \multicolumn{2}{c}{CIFAR100} \\
    \cmidrule(lr){2-3} \cmidrule(lr){4-5}
    \methodname variant & WRN-40-2 & WRN-28-10 & WRN-40-2 & WRN-28-10  \\ 
     \midrule
Uniform policy & 96.22 $\pm$ .10 & 97.38 $\pm$ .05 & 79.07 $\pm$ .24 & 83.26 $\pm$ .17 \\
$\mu$ only & 96.20 $\pm$ .08 & 97.42 $\pm$ .05 & 79.22 $\pm$ .17 & 83.57 $\pm$ .18 \\
$\pi$ only & 96.22 $\pm$ .09 & 97.35 $\pm$ .04 & 79.36 $\pm$ .11 & 83.45 $\pm$ .15 \\
$\pi$ and $\mu$ (ours) & 96.29 $\pm$ .08 & 97.46 $\pm$ .06 & 79.87 $\pm$ .11 & 84.08 $\pm$ .16 \\
\bottomrule
\end{tabular}
}
\caption{CIFAR10/100 accuracy when only learning part of the policy parameters}
\label{tab:norm}
\end{table}
\section{Conclusion}

In this paper, we address the task of automatic data augmentation. Considering the more challenging bilevel optimization problem that arises when the search space is not reduced with default transformations, our proposed \methodname{} method tackles the resulting stability issues thanks to a multi-stage approach based on cold-start, coupled with a KL-regularization. Combined, they allow to reduce the variance of the gradient estimate and to better control the optimization process. We have experimentally observed that our method performs on par with recent approaches leveraging prior knowledge. It has also proved versatile enough to select domain-specific transformations when confronted to non-natural images.

\myparagraph{Acknowledgments} This work was supported by~ANR~3IA MIAI@Grenoble~Alpes~(ANR-19-P3IA-0003) and performed using HPC resources from GENCI–IDRIS (Grant 2022-AD011013343).

{\small
\bibliographystyle{ieee_fullname}
\bibliography{biblio_}
}
\newpage

\appendix

In this supplementary, we first give a short overview of prior work's key aspects (Section~\ref{sec:supp_prior}), then we provide
additional details about the experimental protocol that was used to obtain our results (Section~\ref{exp:supp_exp_setup}) and analyse in more details the results reported in the paper and the impact of our approach on the stability of the search process (Section~\ref{sec:supp_analysis}).

\etocdepthtag.toc{mtappendix}
\etocsettagdepth{mtappendix}{subsection}
\etocsettagdepth{mtchapter}{none}

{
  \hypersetup{linkcolor=ForestGreen}
  \tableofcontents
}

\section{Summary of prior art}
\label{sec:supp_prior}

\begin{table}[t!]
 \scriptsize
 \centering
 \begin{tabular}{llccr}
 \toprule
 Method & Optimization & Prior-free$^*$ & Full set & Search time  \\ 
 \midrule
 AA~\cite{cubuk19autoaugment} & Reinfor. learning & \xmark & \xmark & 5000 \\
 PBA~\cite{ho19population} & Population-based & \xmark & \xmark  & 5 \\
 Fast AA~\cite{lim19fastaa} & Bayesian optim. & \xmark &  \xmark & 3.5 \\ 
 Faster AA~\cite{hataya20fasteraa} & RELAX & \xmark  & \xmark & 0.23  \\
 DADA~\cite{li20dada} & RELAX & \xmark & \xmark & 0.1 \\ 
 RA~\cite{cubuk20ra} & Exhaustive search & \xmark & {\color{SeaGreen}{\cmark}}& 25  \\
 TrivialA \cite{muller21ta} & No optimization & \xmark & {\color{SeaGreen}{\cmark}} & NA \\
 DeepAA~\cite{zheng22deepaa} & Greedy algorithm &  
 {\color{white}{T}}{\color{Orange}{\cmark}}$^{**}$
 & {\color{white}{T}}{\color{white}{T}}{\color{SeaGreen}{\cmark}}$^{***}$ & 9 \\
 \textbf{\methodname (Ours)} & REINFORCE & {\color{SeaGreen}{\cmark}} & {\color{SeaGreen}{\cmark}} &  4 \\ 
 \bottomrule
 \end{tabular}
 \newline
 \scriptsize{\begin{tabular}{l}
 * Prior-free refers to methods that do not use any default transformations.  \\ ** DeepAA does not use default transformations during search but during pretraining.  \\ *** DeepAA pretrains on a reduced dataset.    
 \end{tabular} }
\caption{
\textbf{Key aspects of automatic data augmentation methods. \newline}
 Our approach, \methodname, tackles the corresponding bilevel optimization problem using the REINFORCE gradient estimator. It is prior-free (\ie does not rely on default transformations) and can use the full training set while maintaining a reasonable search time (indicated in GPU hours, for search on CIFAR with WRN-40-2).
 \label{tab:prior}
 }
 \end{table}

Overall, methods for augmentation search
mostly differ in key design choices that are highlighted in Table~\ref{tab:prior}.

\section{Experimental setup}
\label{exp:supp_exp_setup}

In this section, we first describe the practical implementation of our gradient estimation, then we compare our magnitude ranges to those of other methods, and finally we describe in more detail our search and evaluation protocols.

\subsection{Gradient estimation}

In this section, we describe the practical implementation of the gradient estimates derived in our method section.

To optimize the augmentation policies, we minimize an approximation to the upper-level objective $\mathcal{F}(\phi) := \Lval(\theta^{\star}(\phi))$ defined as $\hat{\mathcal{F}}(\phi) := \Lval(\hat{\theta}(\phi))$, where we replaced the intractable lower-level solution $\theta^{\star}(\phi)$ by an approximate solution $\hat{\theta}(\phi)$. Such approximate solution is obtained by performing one gradient step to optimize the lower-level objective starting from the current parameter $\theta$, i.e. $\hat{\theta}(\phi)  := \theta - \eta \nabla_\theta \Ltrain(\theta, \phi)$. 
The gradient $\nabla_{\phi} \mathcal{F}$  is then naturally approximated by $\nabla_{\phi} \hat{\mathcal{F}}$ which is computed by applying the chain rule:
$$\nabla_{\phi} \mathcal{F}\approx  \nabla_{\phi} \hat{\mathcal{F}}= \nabla_{\theta} \Lval(\hat{\theta}(\phi))^{\top} \nabla_\phi\hat{\theta}(\phi).$$ 
The Jacobian $\nabla_{\phi} \hat{\mathcal{F}}$ can be computed explicitly using the Score method which yields:
\begin{align*}
\nabla_\phi \hat{\theta}(\phi) = -\eta
\mathbb{E}_{\tau \sim p_\phi} \left[\nabla_{\theta}\ell_{\text{train}}(\theta,\tau)\nabla_\phi \log p_\phi(\tau)^{\top} \right].
\label{exp-reinforce}
\end{align*}

In practice, expectations over the data and augmentation policies  are estimated with batches. At a given iteration, we sample
$B_{\text{aug}}$ augmentations from $p_{\phi}$ and then apply each of them to a batch of training data $B_{\text{train}}$ from $\Dtrain$ to approximate $\nabla_{\phi}\hat{\theta}(\phi)$. 
Finally, we use a batch $B_{\text{val}}$ of data from $\Dval$ to estimate the validation loss. Denoting $N_a, N_t, N_v$ the size of the augmentation, training and validation batches respectively,
and \begin{align*}
\hat l_{\text{val}}(\theta) &:= \frac{1}{N_v}  \sum_{(x,y)\in B_{\text{val}}} \left[ \ell( y, f_\theta(x)) \right ], \\ \hat l_{\text{train}}(\theta, \tau) &:= \frac{1}{N_t}\sum_{(x,y)\in B_{\text{train}}} \left[ \ell( y, f_\theta( \tau(x))) \right ] 
,
\end{align*}
our gradient estimate can be expressed as
\begin{align*}
    \nabla_{\phi} \mathcal{F} & \approx  -\frac{\eta}{N_a} \nabla_{\hat\theta} \hat l_{\text{val}}(\theta) \left (  \sum_{\tau \in B_{\text{aug}}} \nabla_\theta \hat l_{\text{train}}(\theta, \tau)\nabla_\phi \log p_\phi(\tau)^T \right ) \\
    & = -\frac{\eta}{N_a} \sum_{\tau \in B_{\text{aug}}} \left (  \nabla_{\theta} \hat l_{\text{val}}(\theta)^T \nabla_\theta \hat l_{\text{train}}(\theta, \tau)\right )\nabla_\phi \log p_\phi(\tau) 
\end{align*}
In other words, the upper-level gradient is a weighted sum of the scores $\nabla_\phi \log p_\phi(\tau)$, with the weights representing the alignment between the gradients of i) the loss on the training data transformed with $\tau$ (evaluated at $\theta$), and ii) the loss on the validation data (evaluated at $\hat{\theta}$, \emph{i.e.} one step ahead).

In practice,  the lower-level learning rate decreases with a cosine schedule. As we do not want our upper-level gradient updates to shrink, we set $\eta$ to the \emph{initial} value of the lower-level learning rate instead of its current value.

\subsection{Magnitude ranges.}

The ranges used for mapping the magnitudes to [0,1] vary across methods; we indicate this mapping for each method in Table~\ref{tab:mag-ranges}.
For transformations with respect to which the datasets naturally exhibit symmetries (Shear, Translate, Rotate, Enhance), once we have sampled a magnitude, we randomly select a direction.
Note that \methodname's ranges are larger than the usual ones (\ie those of TA~(RA)), which gives more flexibility during the optimization of our magnitude upper-bounds $\mu$. The latter is initialized at 0.75. Experimentally, we noted that this initialization should be high enough to favour exploration and avoid over-fitting during pre-training. We observed that any initialization in the $[0.75, 0.9]$ range consistently works well across datasets. 

\begin{table*}[t!]
\centering
\footnotesize
\captionsetup{justification=centering}
\begin{tabular}{cccccc}
    \toprule
    \multirow{2}{*}{Application} & \multirow{2}{*}{Transformation} & \multicolumn{4}{c}{Method} \\
      \cmidrule(lr){3-6}
    &  & TA (RA) & TA (Wide) & DomainBed & Ours \\
  \midrule
 \multirow{9}{*}{Sampled} & ShearX/Y & $[0, 0.3]$ & $[0, 0.99]$ & - & $[0, 1]$  \\
   & Translate X/Y & $[0, 0.45]^*$ & $[0, 32\text{px}]^{**}$ & - & $[0, 0.75]$  \\
   & Rotate & $[0, 30]$ & $[0, 135]$ & - & $[0, 90]$ \\
   & Posterize & $[4,8]$ & $[2,8]$ & - & $[2,8]$ \\
   & Solarize & $[0,255]$ & $[0,255]$ & - & $[0, 255]$ \\
   & Enhance$^{***}$ & $[0, 0.9]$ & $[0, 0.99]$ & - & $[0, 0.99]$ \\
   & Cutout & $[0,0.2]$ & $[0,0.6]$ & - & $[0, 1]$ \\
   & RandCrop & - & - & - & $[0, 0.5]$ \\
   & RandResizeCrop & - & - & - & $[0.05, 1]$ \\
    \midrule
    Default & ColorJitter$^{****}$ & $[0, 0.4]$ & $[0,0.4]$ & $[0,0.3]$ & -  \\
    ImageNet/DomainNet & RandResizeCrop & $[0.08, 1]$ & $[0.08, 1]$ & $[0.7, 1]$ & - \\
    \midrule
    \multirow{2}{*}{Default CIFAR} & Cutout & 0.5 & 0.5 & NA & - \\ 
    & RandCrop & 0.125 & 0.125 & NA & - \\
   \bottomrule
\end{tabular}
\caption{Our magnitude ranges compared to those used by other methods. \\
\begin{tabular}{ll}
$^{*}$ TrivialAugment~\cite{muller21ta} uses $[-0.31, 0.31]$, &
$^{**}$ TrivialAugment~\cite{muller21ta} sets the upper-bounds in pixels, not in proportion \\
$^{***}$ Color, Contrast, Brightness, Sharpness, &
$^{****}$ Color, Contrast, Brightness \\
\end{tabular}
}
\label{tab:mag-ranges}
\end{table*}

\subsection{Image pre-processing}
Table \ref{tab:preprocessing} indicates the image pre-processing choices on ImageNet-100 and DomainNet for TrivialAugment~\cite{muller21ta}, DomainBed~\cite{gulrajani2020domainbed} and SLACK.

ImageNet-100 and DomainNet images have variable original sizes. In the literature, training images are commonly resized with RandomResizeCrop. For testing,  TrivialAugment uses Resize(256)+CenterCrop((224,224)), preserving the aspect ratio, while DomainBed directly applies Resize((224,224)), degrading the aspect ratio but preserving the image content.
For each method, we stick to the authors' choices, as we experimentally noted that they yield the best results (\eg using TrivialAugment's pre-processing for DomainBed degrades the performance, and vice-versa). 

For SLACK, which does not apply RandomResizeCrop by default, we preprocess the training data and validation/testing data in the same way. For training, random cropping is applied instead of center cropping to fully exploit the data. For ImageNet-100, we use TrivialAugment's pre-processing. For DomainNet, we select the pre-processing strategy by cross-validation after pre-training.

\begin{table*}[t!]
\footnotesize
\centering
\begin{tabular}{lcccccccc}
\toprule
Dataset & Model & Train & Test \\
\midrule
\multirow{2}{*}{ImageNet-100} & TrivialAugment & RandResizeCrop((224,224)) & Resize(256)+CenterCrop((224,224))  \\
& SLACK & Resize(256)+RandomCro((224,224)) & Resize(256)+CenterCrop((224,224)) \\
\midrule
\multirow{3}{*}{DomainNet} & TrivialAugment ImageNet & RandResizeCrop(224,224) & Resize(256)+CenterCrop(224,224) \\
& TrivialAugment CIFAR & Resize(256)+RandomCrop((224,224),padding=28) & Resize(256)+CenterCrop((224,224)) \\
& DomainBed & RandResizeCrop((224,224)) & Resize((224,224)) \\
& SLACK (Clipart, Sketch, Quickdraw) & Resize((224,224)) & Resize((224,224)) \\
& SLACK (Painting, Infograph, Real) & Resize(256)+RandomCrop((224,224)) & Resize(256)+CenterCrop((224,224)) \\
\bottomrule
\end{tabular}
\caption{Image pre-processing on ImageNet-100 and DomainNet.}
\label{tab:preprocessing}
\end{table*}

\subsection{Policy search}

Hyperparameters used for policy search are indicated in Table \ref{tab:search-hyperparameters}.
They are chosen to satisfy two criteria that we found to be useful for obtaining a successful policy search: i) the validation loss after re-training should be similar (experimentally, slightly lower) to the one obtained after pre-training, and ii) the probability distributions should vary at the same speed for all datasets. 
Our learning rate is 4 times larger for re-training on CIFAR10 than on CIFAR100. We observed that gradients on CIFAR10 are 4 times smaller in norm than those on CIFAR100, and that re-scaling the updates allows satisfying i) and ii) empirically.
For DomainNet, we adapt the number of re-training steps to the dataset size.
A fixed lower-level learning rate for all datasets experimentally satisfies i). We observed that the lower-level gradients differ in scale for each dataset. Satisfying ii) requires re-scaling the KL regularisation and accordingly changing the upper-level learning rate (so that KL weight $\times$ upper-level lr is constant). 

The upper-level learning rate indicated in the Tables is the one used for updating $\pi$. We divide it by $40$ for the optimization of $\mu$ to ensure slower updates for the magnitude parameter which we found to be sensitive to variations (or by $10$ for ablations removing the KL regularization).

\begin{table*}[t!]
\footnotesize
\centering
\begin{tabular}{lcccccccc}
\toprule
Dataset & Network & Re-train iter & Unrolled iter & Batch size & Lower lr & Upper lr & KL weight $\times$ Upper lr \\
\midrule
CIFAR10/100 & WRN-40-2/WRN-28-10 & 1000 & 400 & $8\times 128$ & $0.4$ / $0.1$ & 1 & 0.02  \\
ImageNet-100 & ResNet-18 & 2000 & 800 & $8\times 256$ & 0.1 & 0.5 & 0.005 \\
DomainNet & ResNet-18 & 800 - 1200 &  400 & $8\times 128$ & 0.1 & 0.625 - 1.25 & 0.01 \\
\bottomrule
\end{tabular}
\caption{Hyperparameters used for search on CIFAR, ImageNet-100 and DomainNet.}
\label{tab:search-hyperparameters}
\end{table*}

\subsection{Policy evaluation}

The hyperparameters used for the evaluation phase are indicated in Table~\ref{tab:hyperparameters}. For CIFAR10 and CIFAR100, we use the same hyperparameters as prior work.

\begin{table*}[t!]
\footnotesize
\centering
\begin{tabular}{lccccc}
\toprule
Dataset & Network & Epochs & Batch size & Learning rate & Weight decay \\
\midrule
CIFAR10/100 & WRN-40-2, WRN-28x10 & 200 & 128 & 0.1 & 0.0005$^*$ \\
ImageNet-100 & ResNet-18 & 270 & 256 & 0.1 & 0.001 \\
DomainNet & ResNet-18 & 200 & 128 & 0.1 & 0.001 \\
\bottomrule
\end{tabular}
\caption{Hyperparameters used for training on CIFAR, ImageNet-100 and DomainNet. \footnotesize{$^*$: 0.0002 for CIFAR10 on WRN-40-2}}
\label{tab:hyperparameters}
\end{table*}

\section{Extended analysis}
\label{sec:supp_analysis}

In this section, we further analyse the results of our search algorithm. We first illustrate the policies learned for all datasets. We then study in more detail the impact of our multi-stage approach with KL regularization, comparing it with single-stage (unrolled) and unregularized approaches and illustrating their instability.

\subsection{Uniform distribution: ablations on prior work}

\begin{table*}[t!]
\centering
\footnotesize
\begin{tabular}{lccllll}
\toprule
\multirow{2}{*}{Model}  & \multirow{2}{*}{Magnitude model} & \multicolumn{2}{c}{CIFAR10} & \multicolumn{2}{c}{CIFAR100} \\
 \cmidrule(lr){3-4} \cmidrule(lr){5-6}
&  & WRN-40-2 & WRN-28-10 & WRN-40-2 & WRN-28-10 \\
\midrule
\multirow{2}{*}{FastAA/DADA initialization} & Theirs & 96.22 & 97.08 & 78.26 & 82.17 \\
& Uniform & 96.37 & 97.25 & 79.10 & 82.80 \\
\midrule
FastAA, reported & Original & 96.4 & 97.3 & 79.3 & 82.7 \\
\cmidrule(lr){1-1}
FastAA, reproduced & Original & 96.4 & 97.22 & 79.11 & 82.82 \\
(evaluation only) & Uniform & 96.37 & 97.30 & 79.15 & 82.84 \\
\midrule
DADA, reported & Original  & 96.4 & 97.3 & 79.1 & 82.5 \\
 \cmidrule(lr){1-1} 
DADA, reproduced & Original & 96.33 & 97.19 & 79.07 & 82.05  \\
(evaluation only) & Uniform & 96.37 & 97.35 & 78.97 & 82.57 \\
\midrule
DeepAA, reported & Original & - & 97.56 & - & 84.02 \\
 \cmidrule(lr){1-1}
DeepAA, reproduced &  Original & 96.46 & 97.48 & 79.62 & 83.85 \\
(evaluation only) & Uniform & 96.55 & 97.47 & 78.89 & 83.62 \\
\bottomrule
\end{tabular}
\caption{\textbf{Why learning the magnitude range?} CIFAR10/100 accuracies for DADA, FastAA and DeepAA, when using their original magnitude model, or a simpler one which samples in a uniform manner in their ranges.} 
\label{tab:magnitude-ablations}
\end{table*}

In this section, we motivate our choice of a uniform magnitude distribution, showing that it globally outperforms optimized magnitude models in prior work. To this end, we directly evaluate the policies provided by the authors without re-running their search procedure. We compare their learned magnitude model with a simpler one that consists in sampling the magnitudes uniformly on their $[0,1]$-mapped ranges.
We study three baselines: DADA~\cite{li20dada}, FastAA~\cite{lim19fastaa} and DeepAA~\cite{zheng22deepaa}. Note that their policy results from a search on CIFAR10, that they also use when evaluating on CIFAR100. Results are reported in Table \ref{tab:magnitude-ablations}. 

\myparagraph{Parametrization} 
DADA and FastAA directly optimize a probability distribution over the set of all possible composite transformations (\emph{sub-policies}) and lean a single magnitude value for each transformation in a sub-policy. They keep the top-$k$ sub-policies for evaluation.  DeepAA learns to compose transformations in a greedy manner and discretizes the magnitude ranges, learning a probability for each magnitude. 
We compare these learned magnitude values (FastAA, DADA) or learned probabilities (DeepAA) to our approach based on a uniform sampling. 

\myparagraph{DADA/FastAA} 
Our approach compares favorably to DADA's and FastAA's optimized models.
We also compare both approaches on their initial policy (equal probabilities for all sub-policies, magnitudes set at mid-range). With uniform magnitude sampling, their initial policy (sampling among all possible sub-policies) performs similarly if not better than their optimized one (sampling among their top-$k$ sub-policies). 

\myparagraph{DeepAA} 
Results on the policy provided by DeepAA are more nuanced: using uniform sampling improves results on CIFAR10 (on which their search was conduced) and degrades them on CIFAR100. 

\subsection{Visualization of the learned policies}

\begin{figure*}[t!]
    \centering
    \begin{subfigure}{0.49\linewidth}      
    \includegraphics[width=0.44\linewidth]{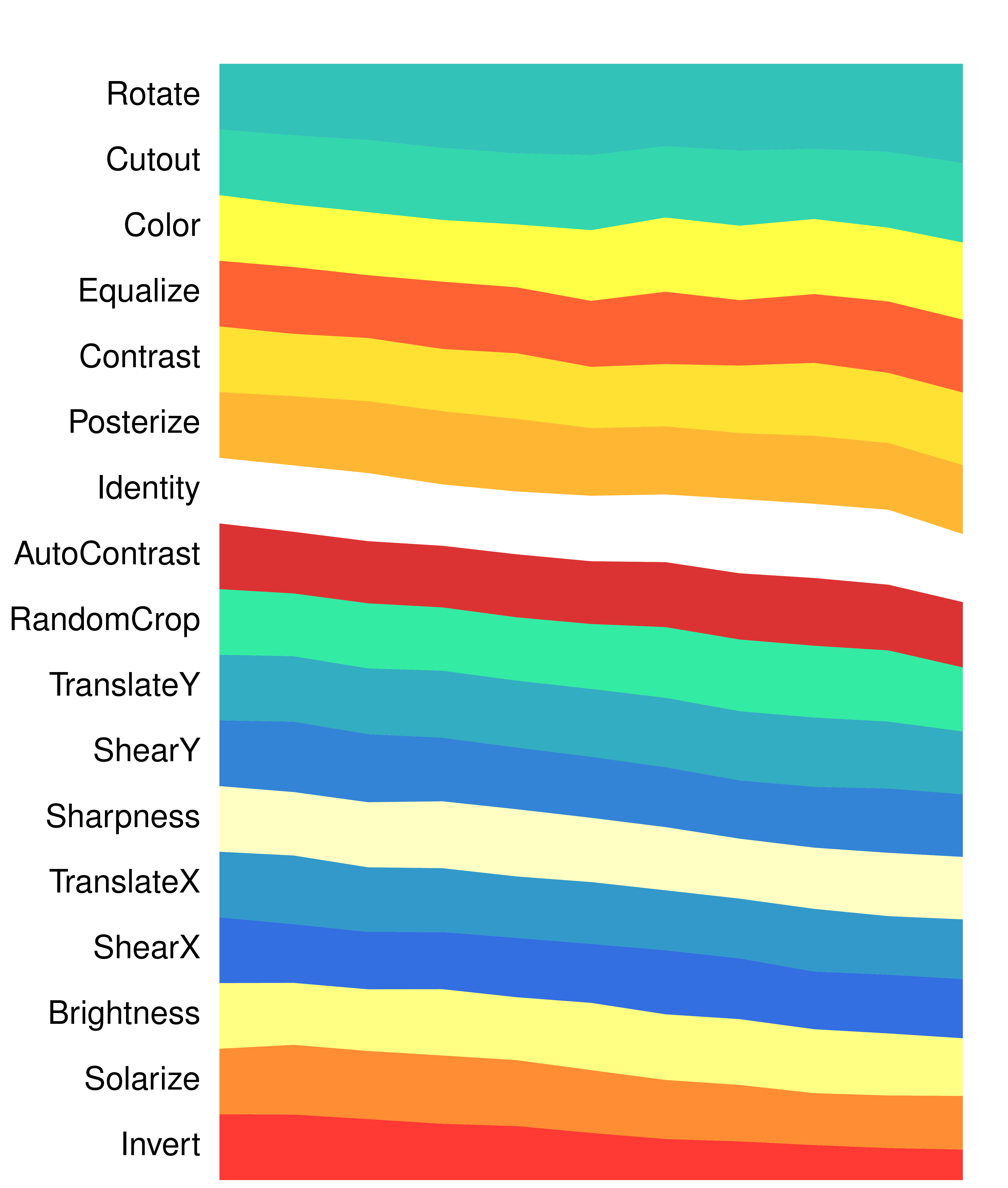}
    \includegraphics[width=0.55\linewidth]{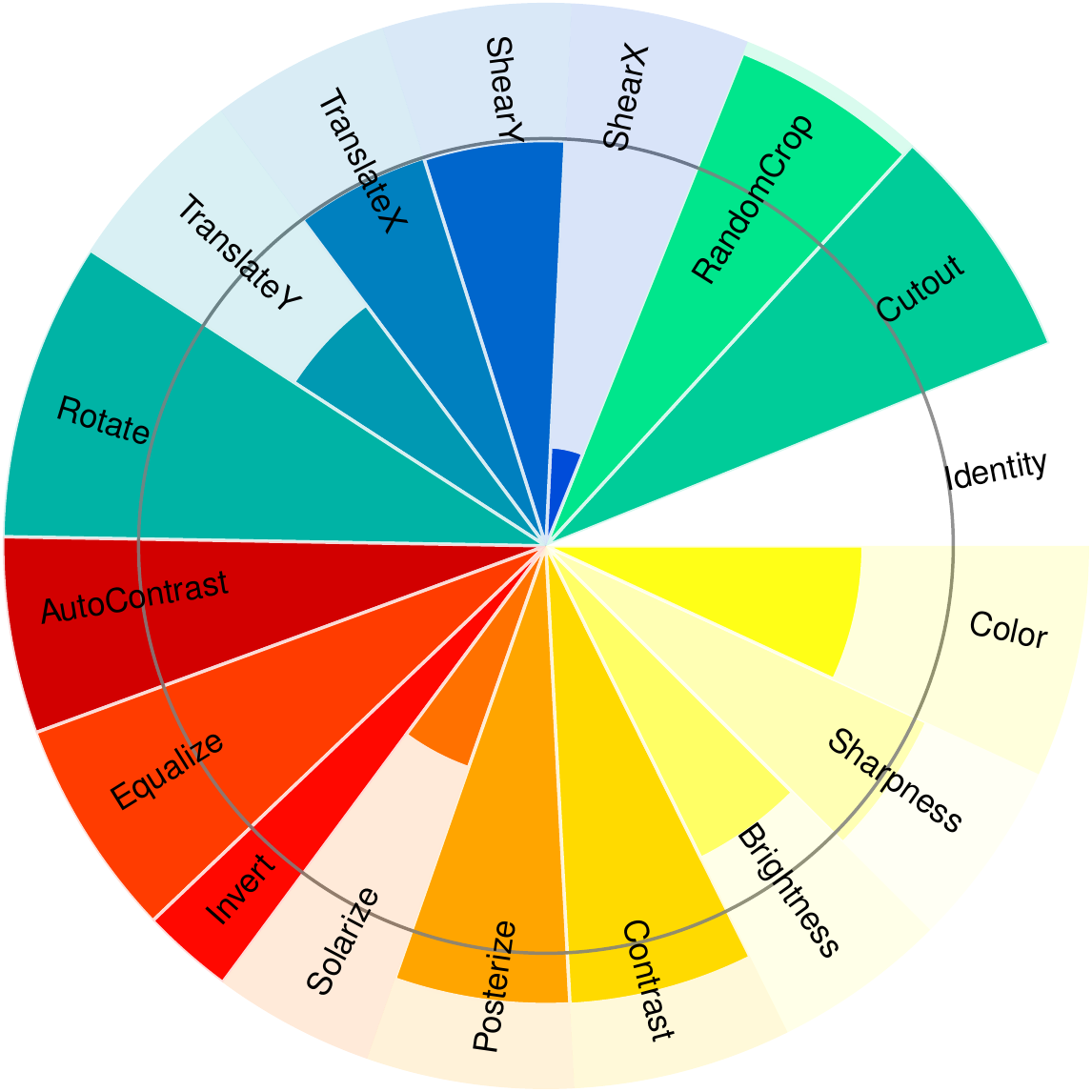}
    \subcaption{CIFAR10, WRN-40x2}      
    \end{subfigure} 
    \begin{subfigure}{0.49\linewidth} 
    \includegraphics[width=0.44\linewidth]{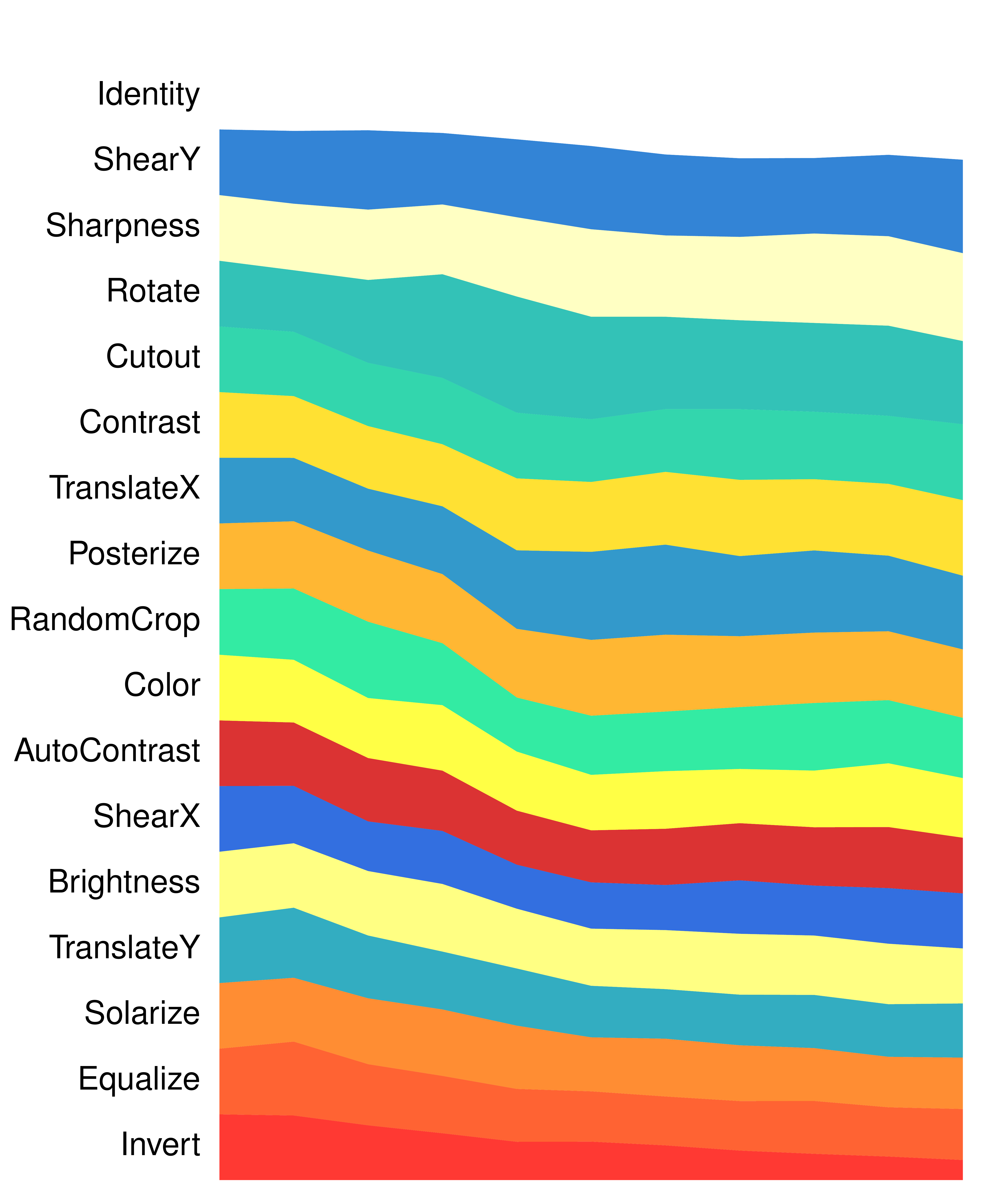}
    \includegraphics[width=0.55\linewidth]{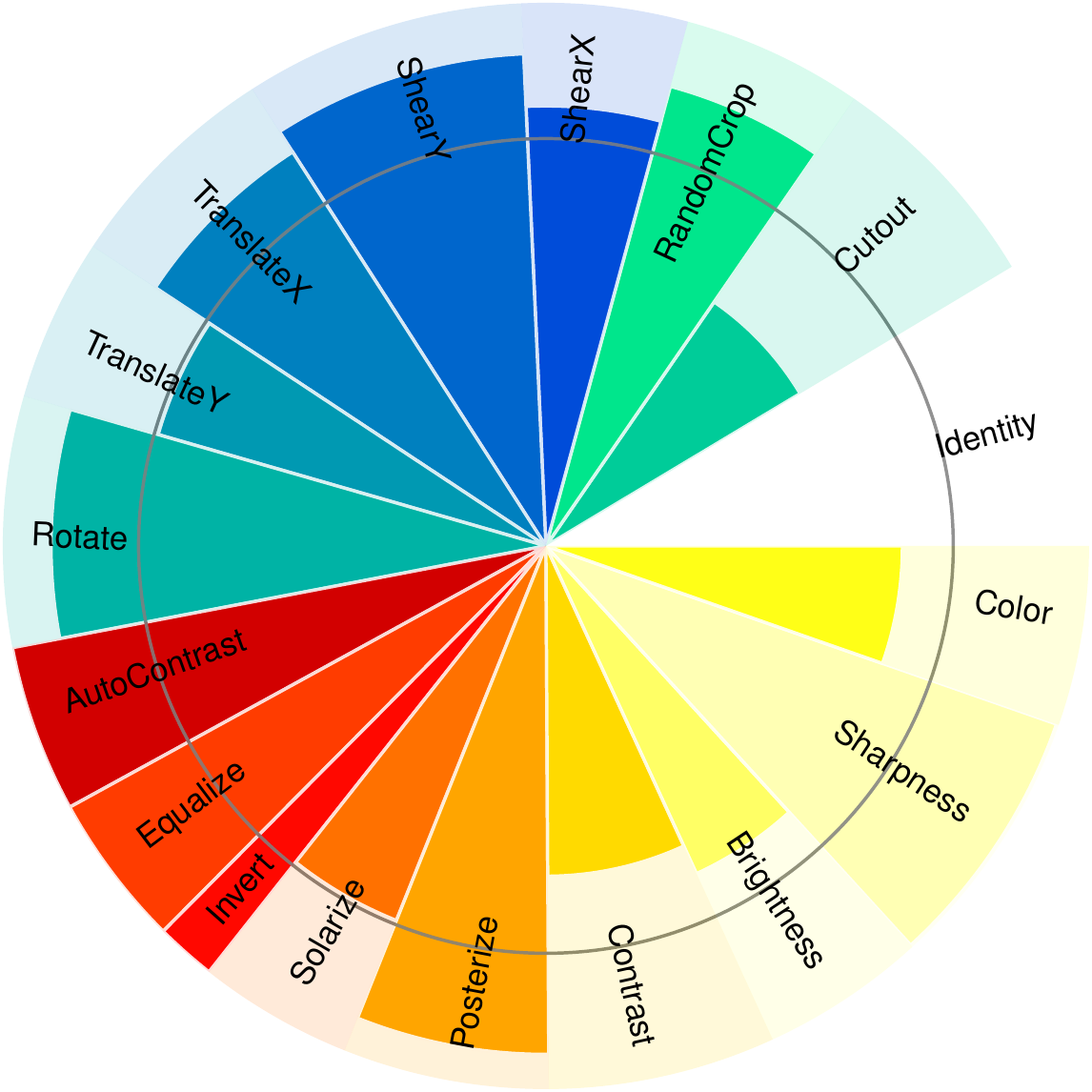}
    \subcaption{CIFAR10, WRN-28x19}      
    \end{subfigure} 
    \begin{subfigure}{0.49\linewidth}      
    \includegraphics[width=0.44\linewidth]{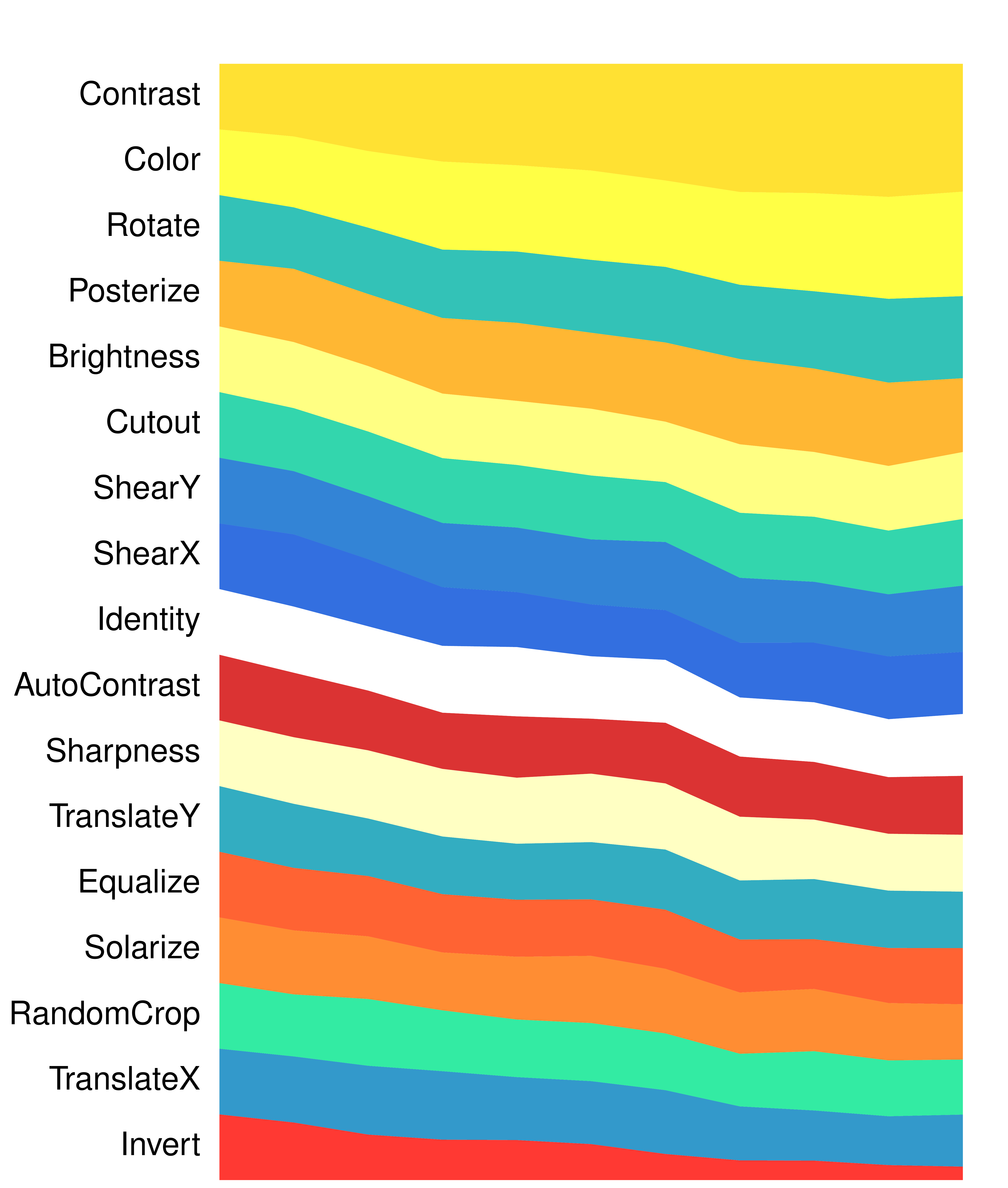}
    \includegraphics[width=0.55\linewidth]{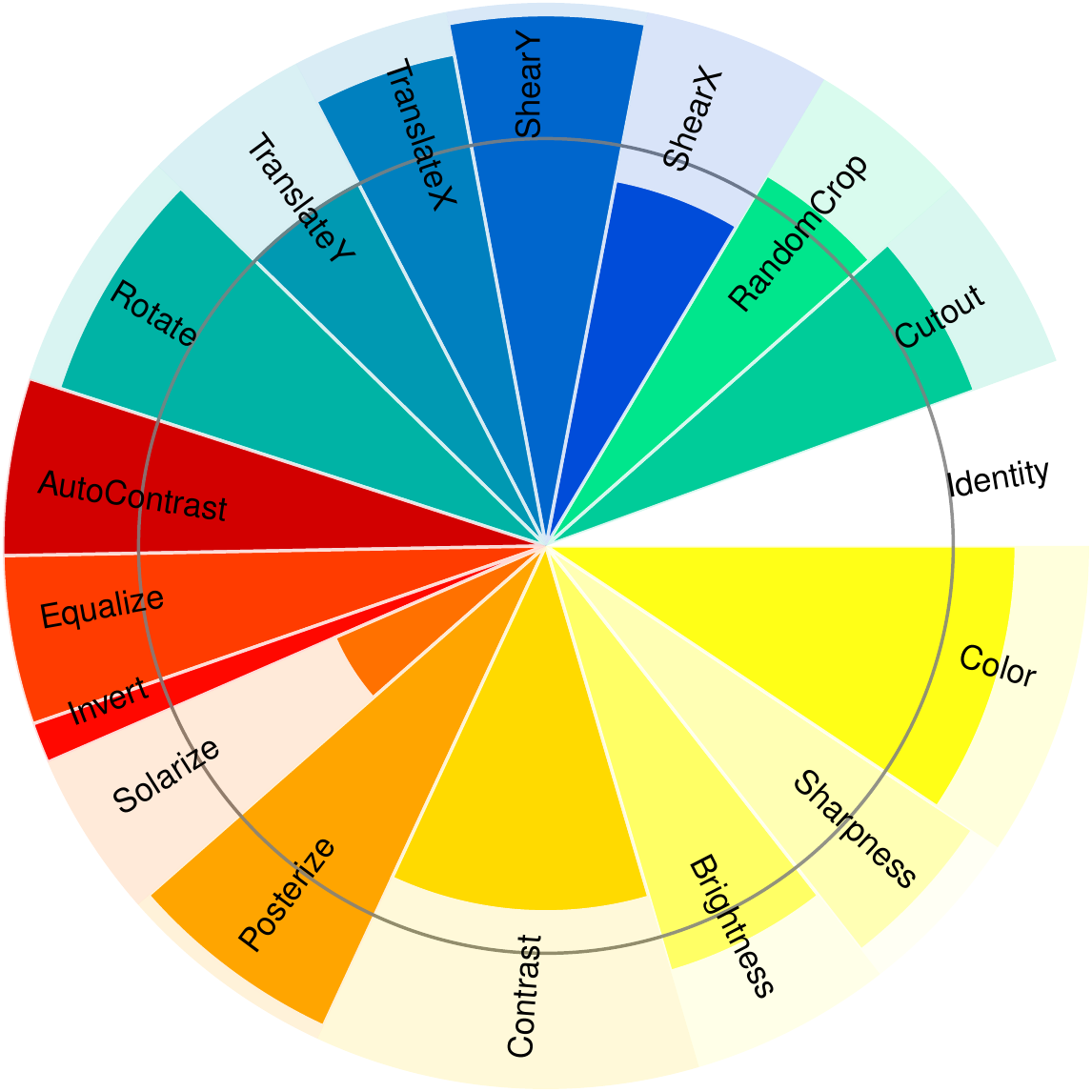}
    \subcaption{CIFAR100, WRN-40x2}      
    \end{subfigure} 
    \begin{subfigure}{0.49\linewidth}
    \includegraphics[width=0.44\linewidth]{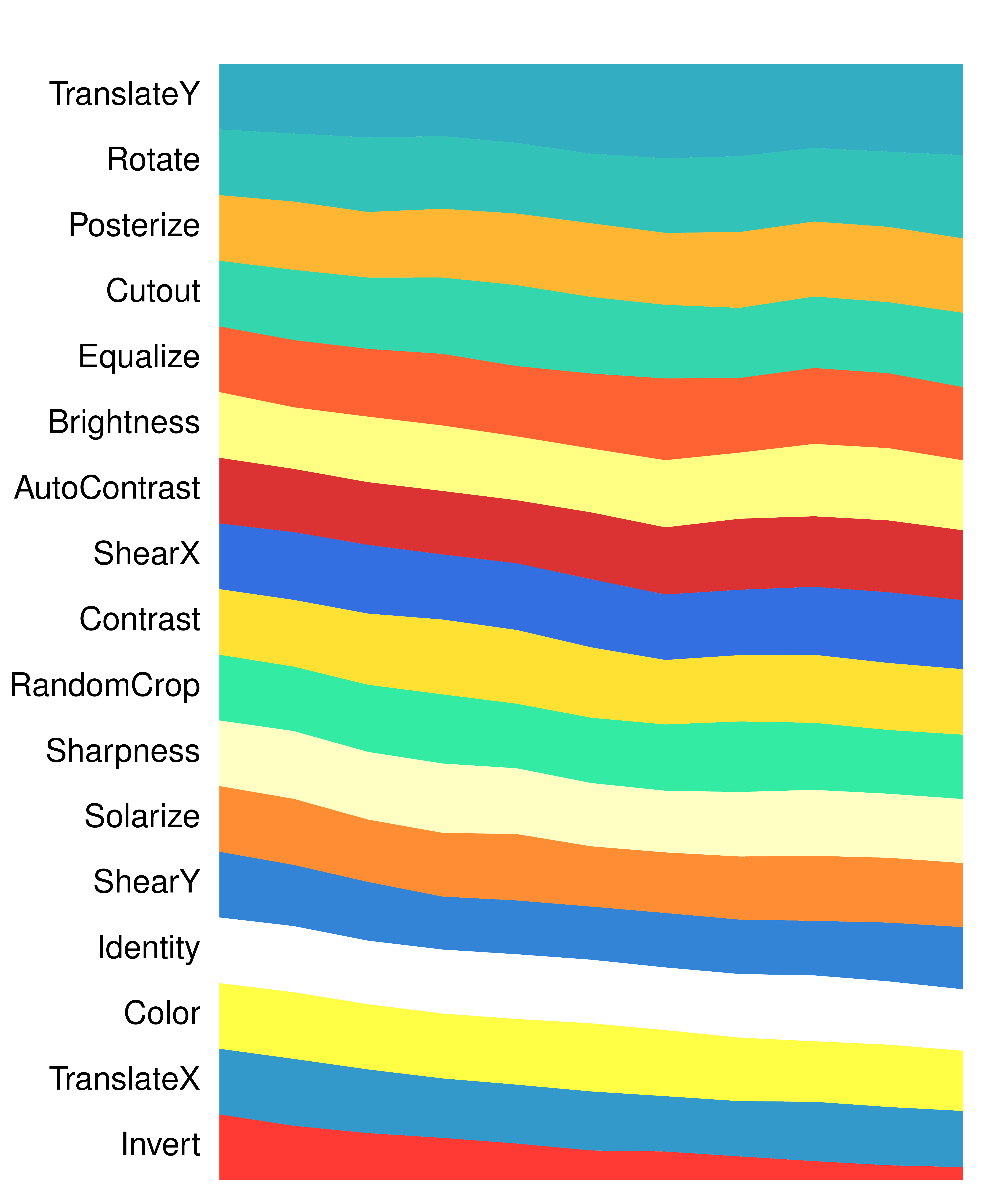}
    \includegraphics[width=0.55\linewidth]{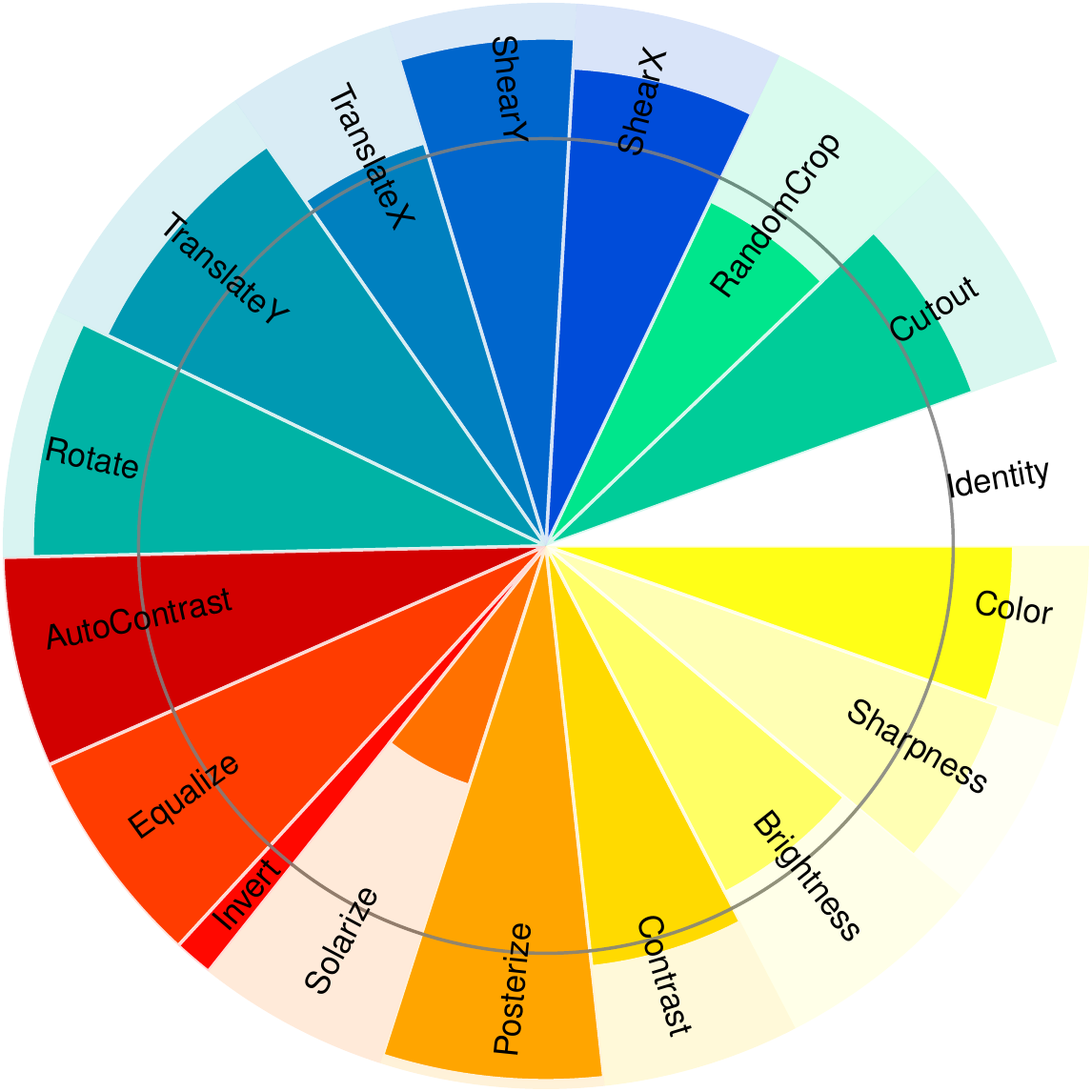}
    \subcaption{CIFAR100, WRN-28x10}      
    \end{subfigure}
    \caption{Illustrations of best policies found for CIFAR10 and CIFAR100 on WideResNet-40x2 and WideResNet-28x10 architectures. For each dataset and architecture, we show the evolution of the probability distribution $\pi$ as training progresses (left) and the final learned policy as a pie chart (right), where slice widths represent $\pi$ and and slice radii represent $\mu$.}
    \label{fig:cifar-pies}
\end{figure*}

\myparagraph{CIFAR}
The evolution of probability distributions for CIFAR10 and CIFAR100 and pie charts of the final policies are illustrated in Fig.~\ref{fig:cifar-pies}. It can be noted that Invert and Solarize, known to be detrimental, are systematically discarded.
The policies learned are quite diverse, 
with different leading transformations for each distribution but global predominance of some transformations such as Cutout or Rotate. 
It can also be noted that magnitudes upper-bounds are in average higher for the larger WideResNet-28x10 networks: a larger learning capacity benefits more from harder transformations.

\begin{figure}[t!]
    \centering
    \includegraphics[width=0.46\linewidth]{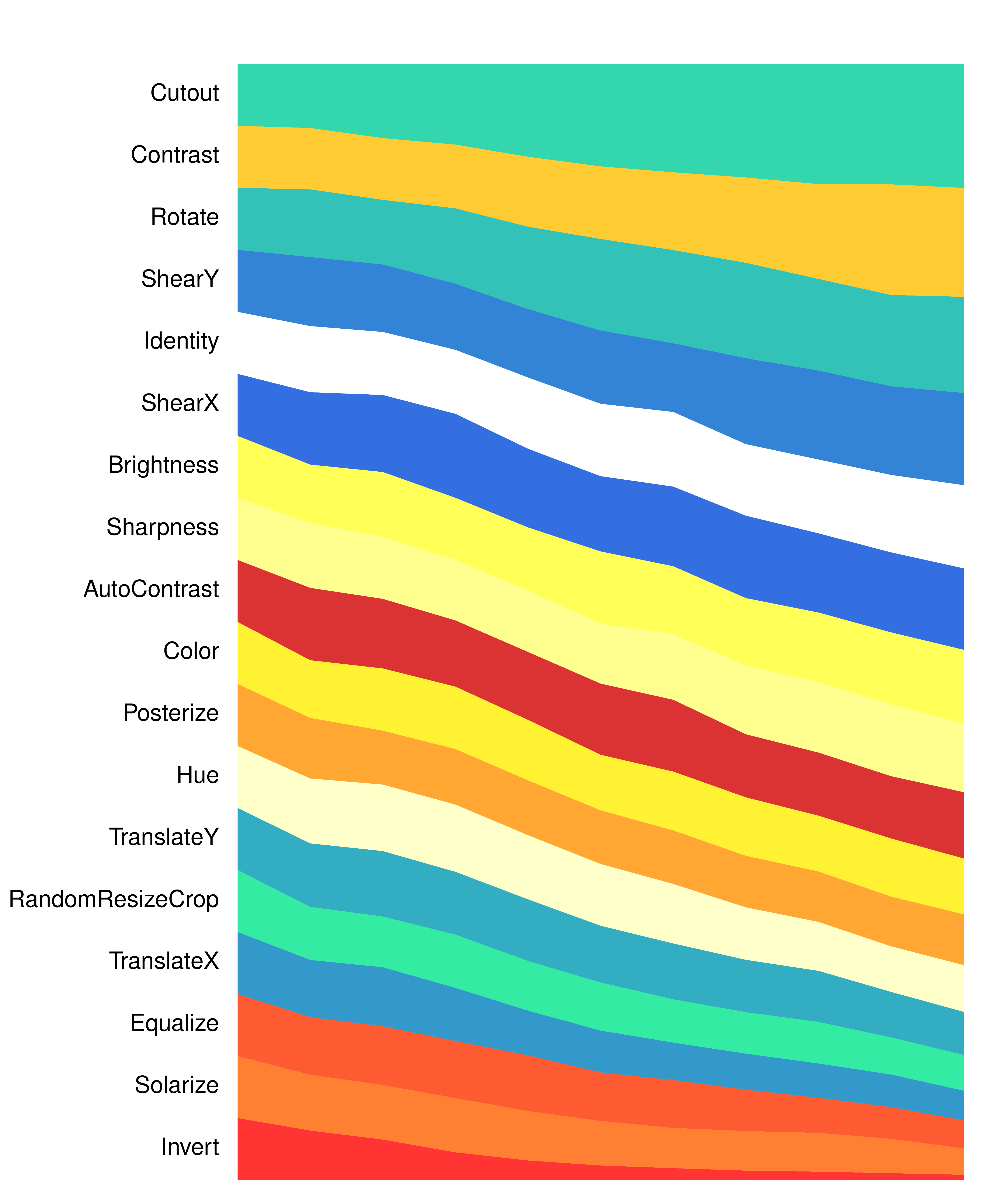}
    \includegraphics[width=0.53\linewidth]{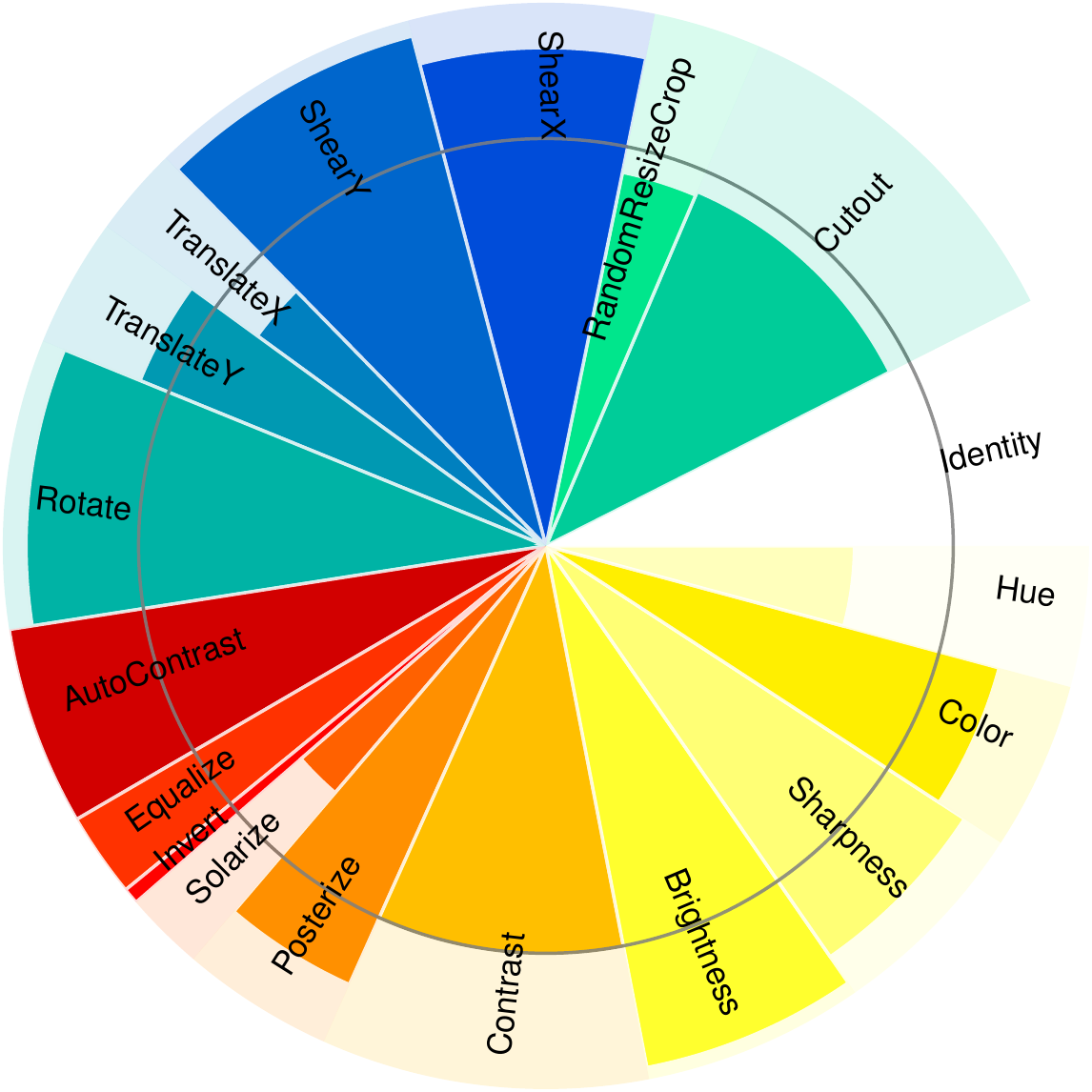}
    \caption{ImageNet-100 policy on the best search split}   
    \label{fig:imagenet100-pies}
\end{figure}

\myparagraph{ImageNet-100}
The best policy found for ImageNet-100 is illustrated in Fig.~\ref{fig:imagenet100-pies}. Interestingly, RandomResizeCrop is ranked quite low, yet our policy yields results comparable to TrivialAugment's (with a $86.18$ average accuracy on this split), suggesting that other geometrical transformations such as Cutout, Rotate and ShearY are equivalently beneficial for training on ImageNet-100. We can note rather high magnitudes upper-bounds for the color jittering transformation TrivialAugment also applies by default (Color, Contrast, Brightness), which is consistent with the higher performance of TrivialAugment's (Wide) version compared to (RA).

\begin{figure*}[t!]
    \centering
    \begin{subfigure}{0.33\linewidth}      \includegraphics[width=\linewidth]{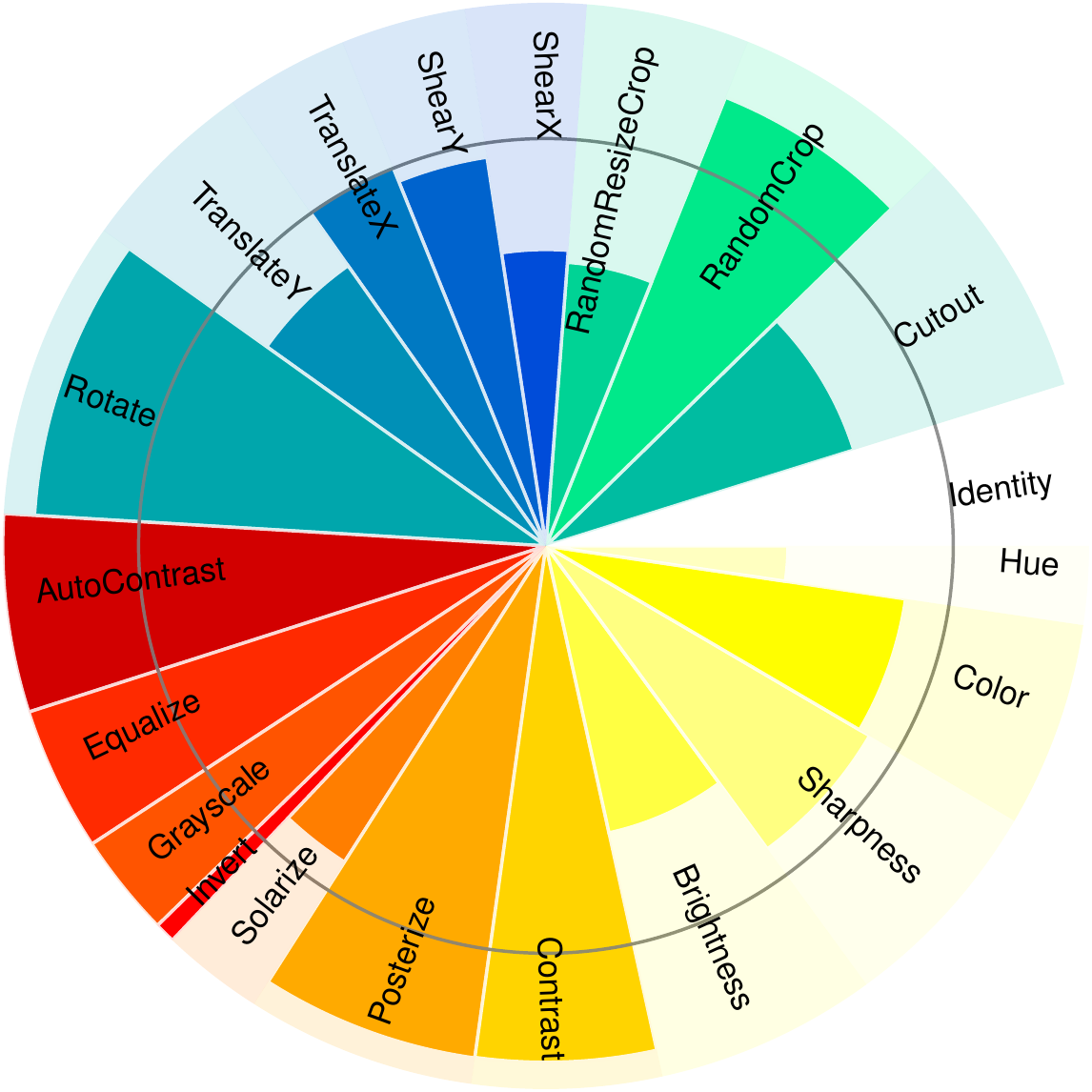}
    \subcaption{Real-50k}      
    \end{subfigure}
        \begin{subfigure}{0.33\linewidth}      \includegraphics[width=\linewidth]{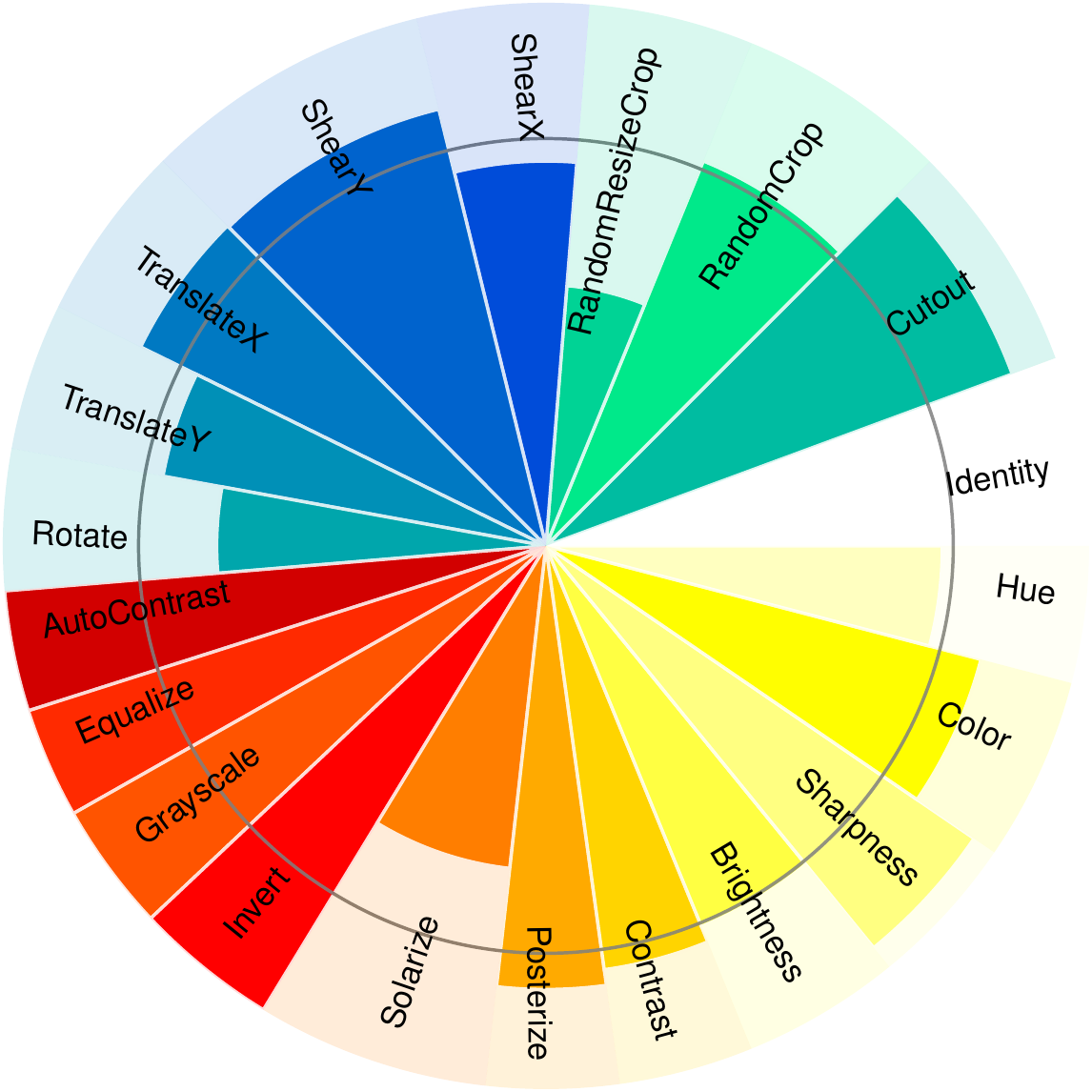}
    \subcaption{Quickdraw-50k}      
    \end{subfigure}
        \begin{subfigure}{0.33\linewidth}      \includegraphics[width=\linewidth]{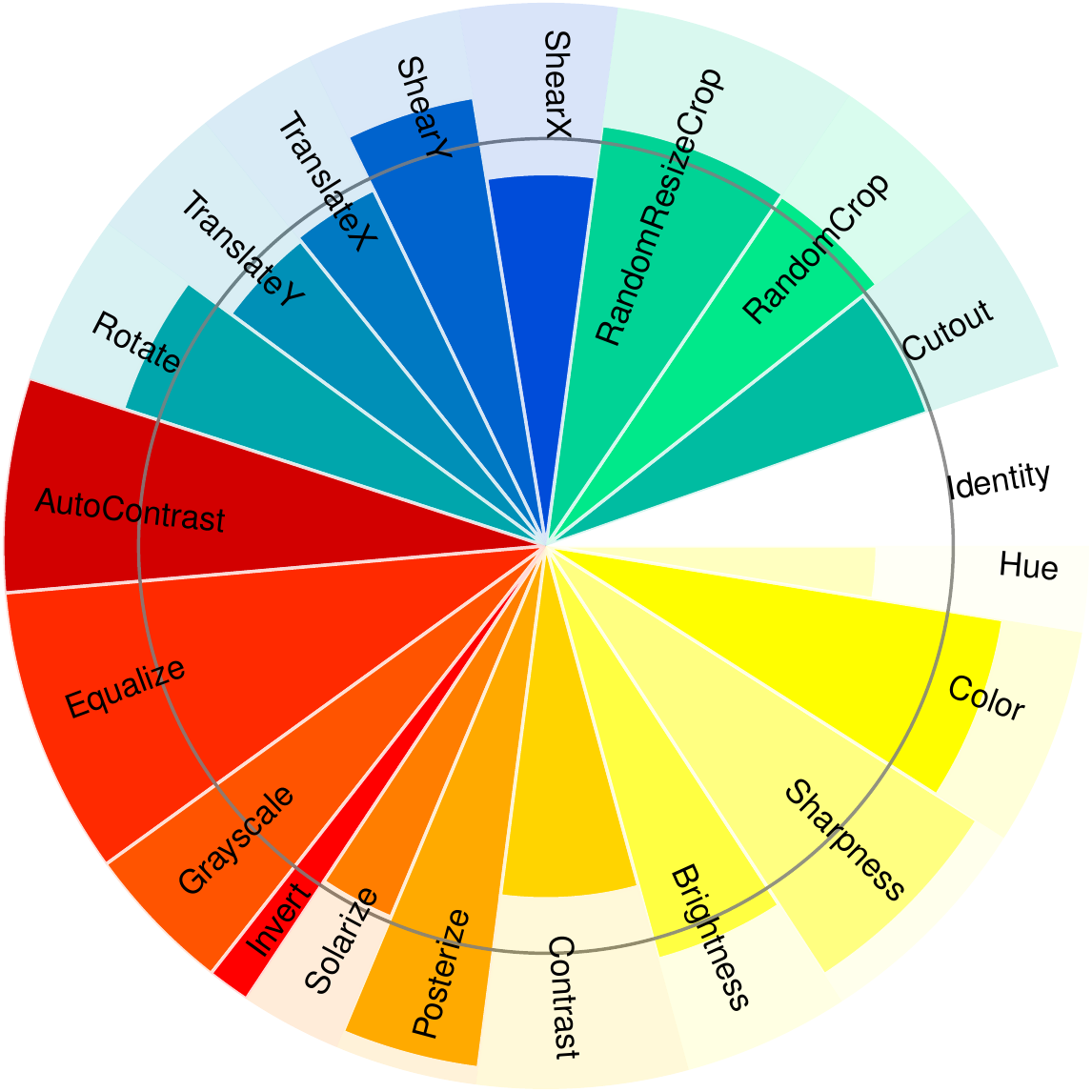}
    \subcaption{Infograph}      
    \end{subfigure}
        \begin{subfigure}{0.33\linewidth}      \includegraphics[width=\linewidth]{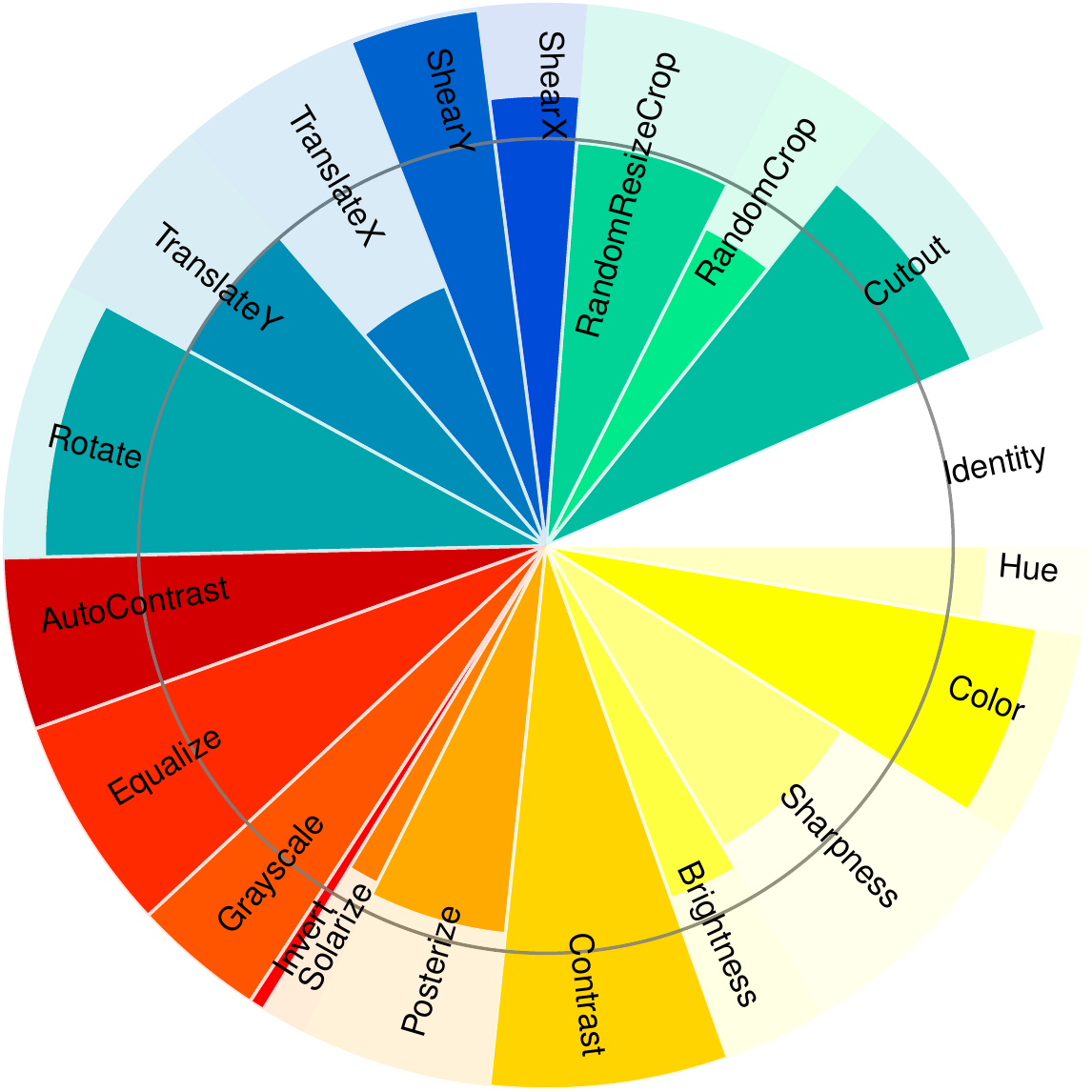}
    \subcaption{Painting}      
    \end{subfigure}
    \begin{subfigure}{0.33\linewidth}      \includegraphics[width=\linewidth]{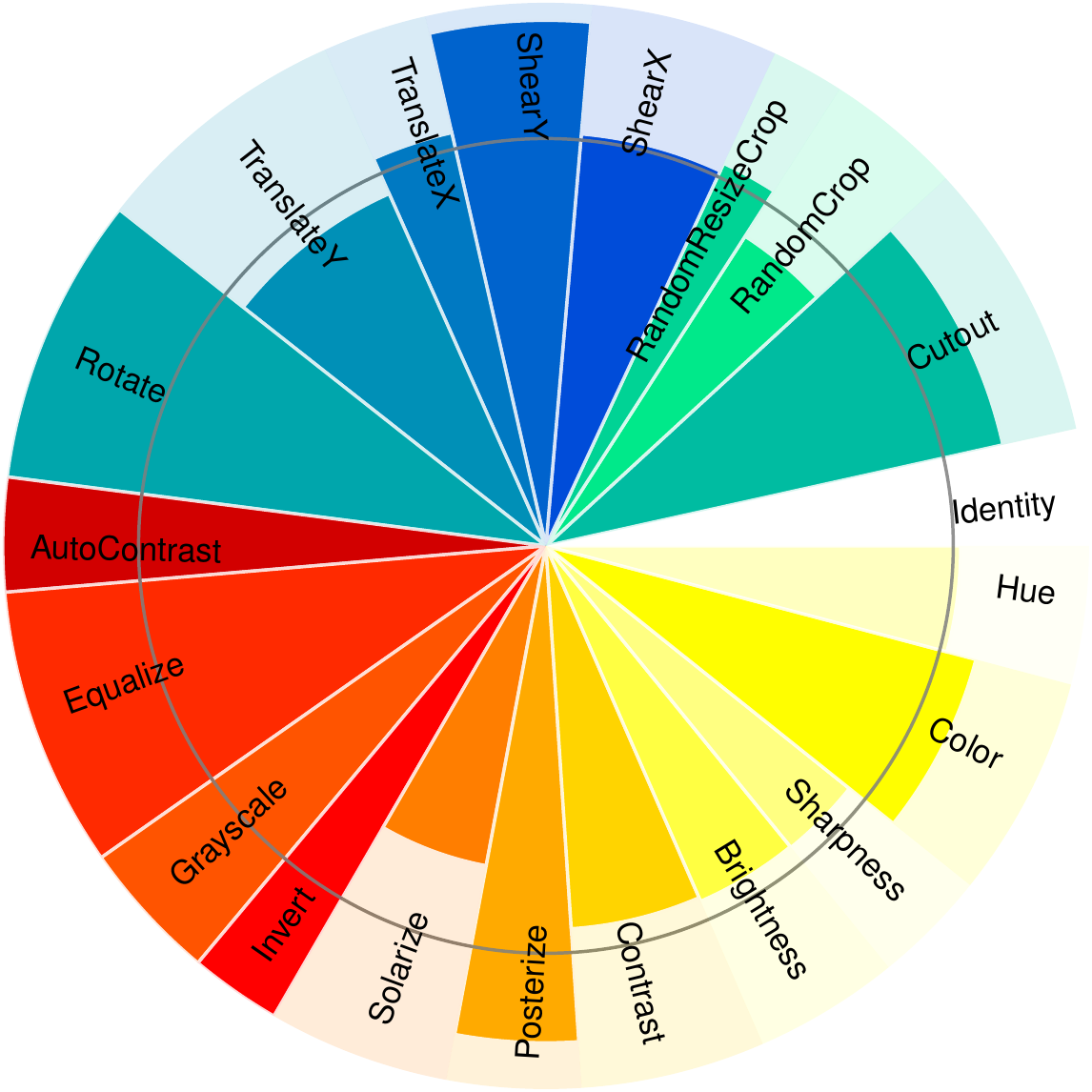}
    \subcaption{Sketch}      
    \end{subfigure}
    \begin{subfigure}{0.33\linewidth}      \includegraphics[width=\linewidth]{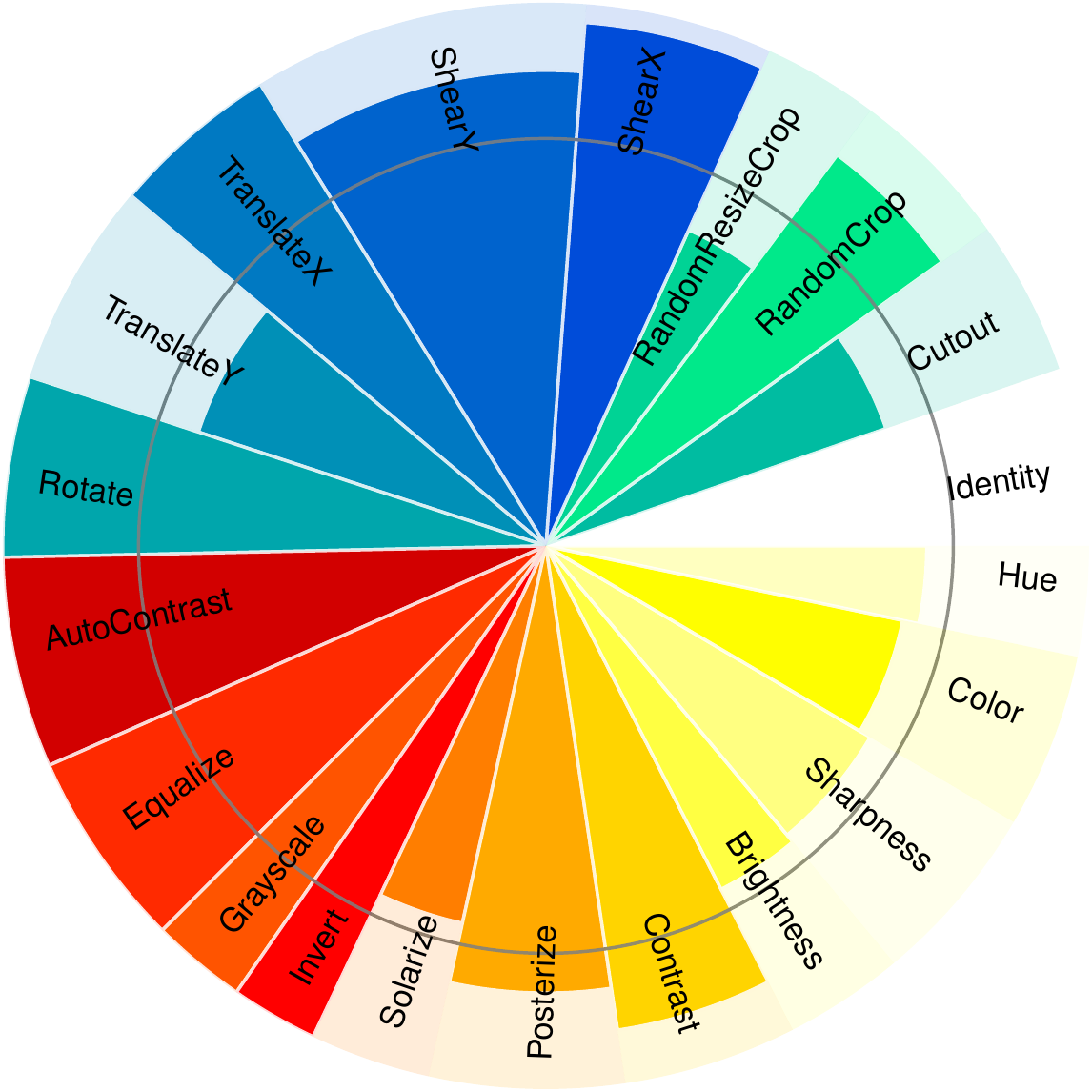}
    \subcaption{Clipart}      
    \end{subfigure}
    \caption{Policies found on DomainNet. The three distributions from $\pi$ forming the composite transformation are averaged.}
    \label{fig:domainnet-pies}
\end{figure*}

\myparagraph{DomainNet}
The policies found by SLACK on DomainNet are illustrated in the pie charts Fig.~\ref{fig:domainnet-pies} for all domains. 
Some similarities with policies found on CIFAR and ImageNet can be noted. In particular, Invert and Solarize (that only inverts part of the pixels) are systematically discarded for all domains except Quickdraw. Invert is manually removed from TrivialAugment's baseline as it is known to be detrimental, and this seems to generalize to other domains. Also, Rotate and Cutout are globally favoured, similarly to the policies found on CIFAR and ImageNet-100. 

However some differences mark specificities to each domain: i) on the strength of the transformations: for example, geometrical transformations are given high magnitudes on Clipart and lower ones on Real, ii) on their probabilities: color jittering transformations used for real images are globally assigned a high probability for Real, Painting and Infograph domains, and a much lower one for Clipart, which suggests that changes in color, contrast or brightness are less meaningful for this domain.

\subsection{Avoiding instabilities with SLACK}

In this section, we study how our augmentation policies evolve when removing the KL regularization or when using a single optimization stage instead of multiple ones, which corresponds to standard unrolled optimization. Evaluations in both settings are reported in Table \ref{tab:ablations}.

We first show that unrolled optimization is globally unstable and easily collapses, justifying the need for a regularization. We illustrate how entropy regularization prevents collapse and yields competitive results, but at the cost of high `local' instability. These instabilities make the final performance highly dependent on the choice of some hyperparameters, such as the learning rate.
The necessity to overcome these instabilities motivates our multi-stage procedure with an adaptive anchoring for the regularization. Lastly, we show that unregularized multi-stage optimization, while more stable than unregularized unrolled optimization, does not yield competitive results, confirming again the benefits of our KL regularization.

\begin{table*}[t!]
  \centering
  \footnotesize
  \begin{tabular}{lccccccc}
     \toprule
    &&&& \multicolumn{2}{c}{CIFAR10} & \multicolumn{2}{c}{CIFAR100} \\
    \cmidrule(lr){5-6} \cmidrule(lr){7-8}
    \methodname variant &  Upper-level iterations  & Upper-level lr & KL weight & WRN-40-2 & WRN-28-10 & WRN-40-2 & WRN-28-10  \\ 
    \midrule
    Unrolled w/ KL (Fig.~\ref{fig:ev-unrolled}) & 10000 & 0.25 & 0.005 & 96.30 $\pm$ .08 & 97.43 $\pm$ .04 & 79.54 $\pm$ .20 & 84.11 $\pm$ .13 \\
    \methodname w/o KL (Fig.~\ref{fig:ev-nkl}) & $10\times 400$ & 0.25 & 0 & 96.27 $\pm$ .05  & 97.06 $\pm$ .11 & 79.61 $\pm$ .13 & 83.79 $\pm$ .19 \\
    \midrule
    \methodname (Fig.~\ref{fig:cifar-pies}) & $10\times 400$ & 1 & 0.02 & 96.29 $\pm$ .08 & 97.46 $\pm$ .06 & 79.87 $\pm$ .11 & 84.08 $\pm$ .16 \\
    \bottomrule
\end{tabular} \\
\caption{CIFAR10/100 accuracy with unregularized and single-stage approaches.}
\label{tab:ablations}
\end{table*}

\begin{figure*}[t!]
    \centering  
    \begin{subfigure}{0.72\linewidth}      \includegraphics[width=\linewidth]{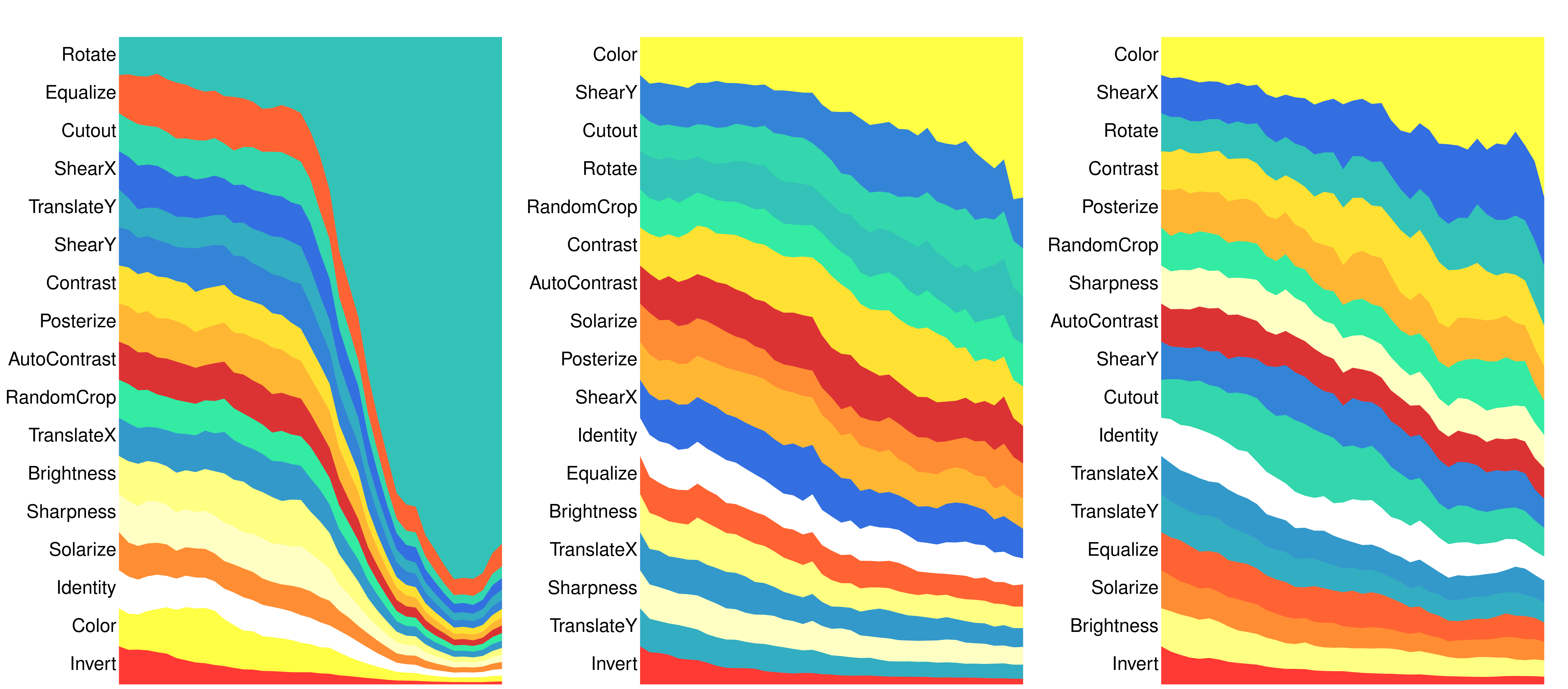}
    \subcaption{The three distributions over transformations forming the composition}      
    \end{subfigure}
    \begin{subfigure}{0.27\linewidth}      \includegraphics[width=\linewidth]{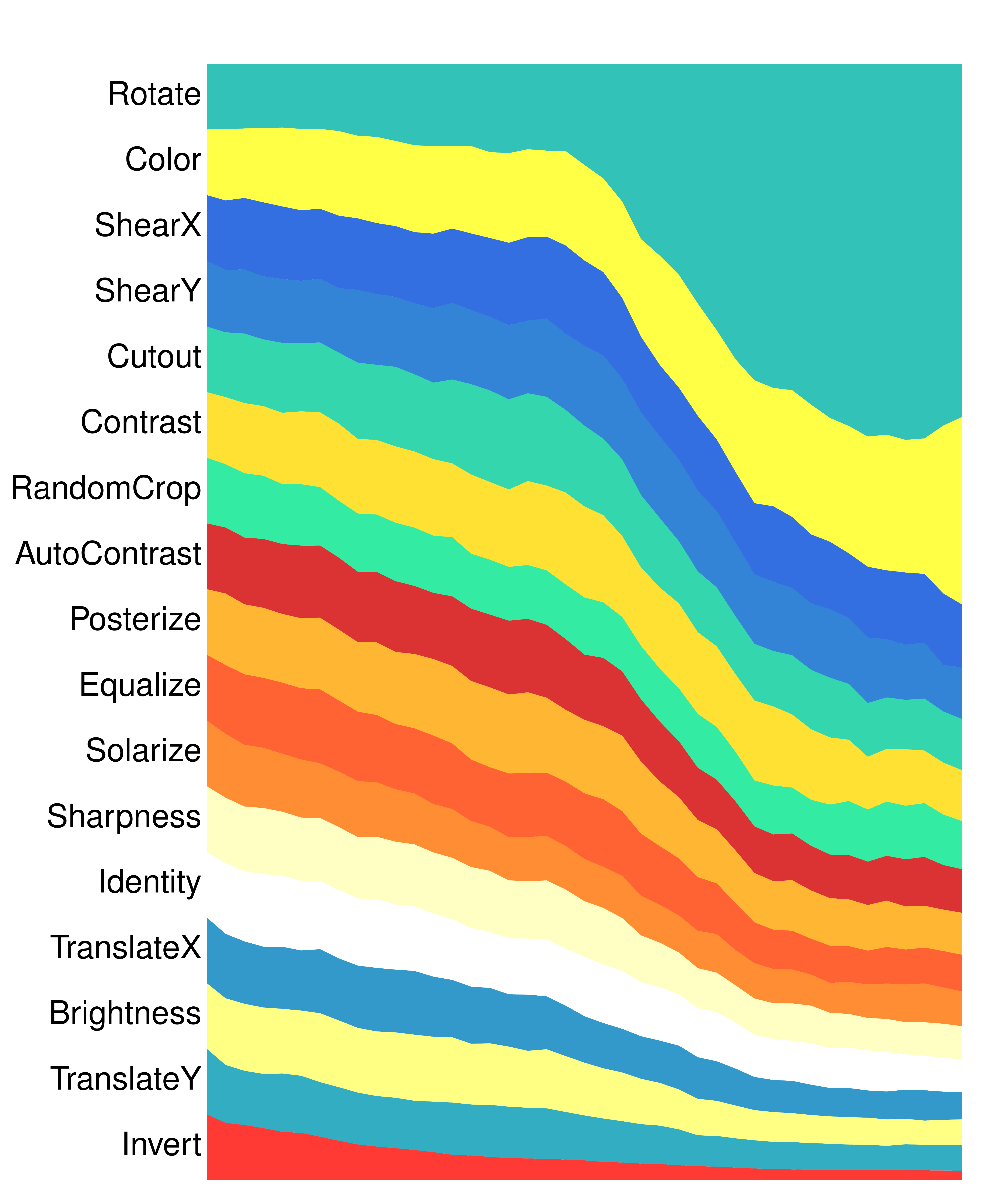}
    \subcaption{Average of the 3 distributions}
    \end{subfigure} 
    \caption{Evolution of the probability distributions $\pi$ for CIFAR100 with unregularized unrolled optimization in a case of collapse.
    }
    \label{fig:ev-unrolled-nkl}
\end{figure*}

\myparagraph{Unregularized unrolled optimization}
Unrolled optimization is subject to two sources of instability: first, the approximation $\theta^\star(\phi) = \hat{\theta}(\phi)$ with a single gradient step inherently leads to wrong gradient updates; second, the REINFORCE gradient estimation is theoretically exact but has a high variance in practice when approximated in the context of stochastic optimization. Fig. \ref{fig:ev-unrolled-nkl} illustrates these instabilities: blindly following wrong gradient directions exacerbated by an oversampling of the dominant transformation leads to a progressive collapse of the policy.

\myparagraph{Unrolled optimization with entropy regularization}
In the case of a single-stage unrolled optimization, the KL regularization uses a uniform distribution as an anchor, which corresponds to an entropy regularization. By maximizing the entropy, the algorithm encourages exploration of the augmentation policies and prevents the divergence phenomenon observed above. While this regularization leads to competitive results as reported in Table \ref{tab:ablations}, it does not mitigate the inherent instability of the gradient updates. On the other hand, the multi-stage algorithm we proposed in SLACK yields more stable gradient updates.

\begin{figure*}[t!]
    \centering
    \begin{subfigure}{0.72\linewidth}      \includegraphics[width=\linewidth]{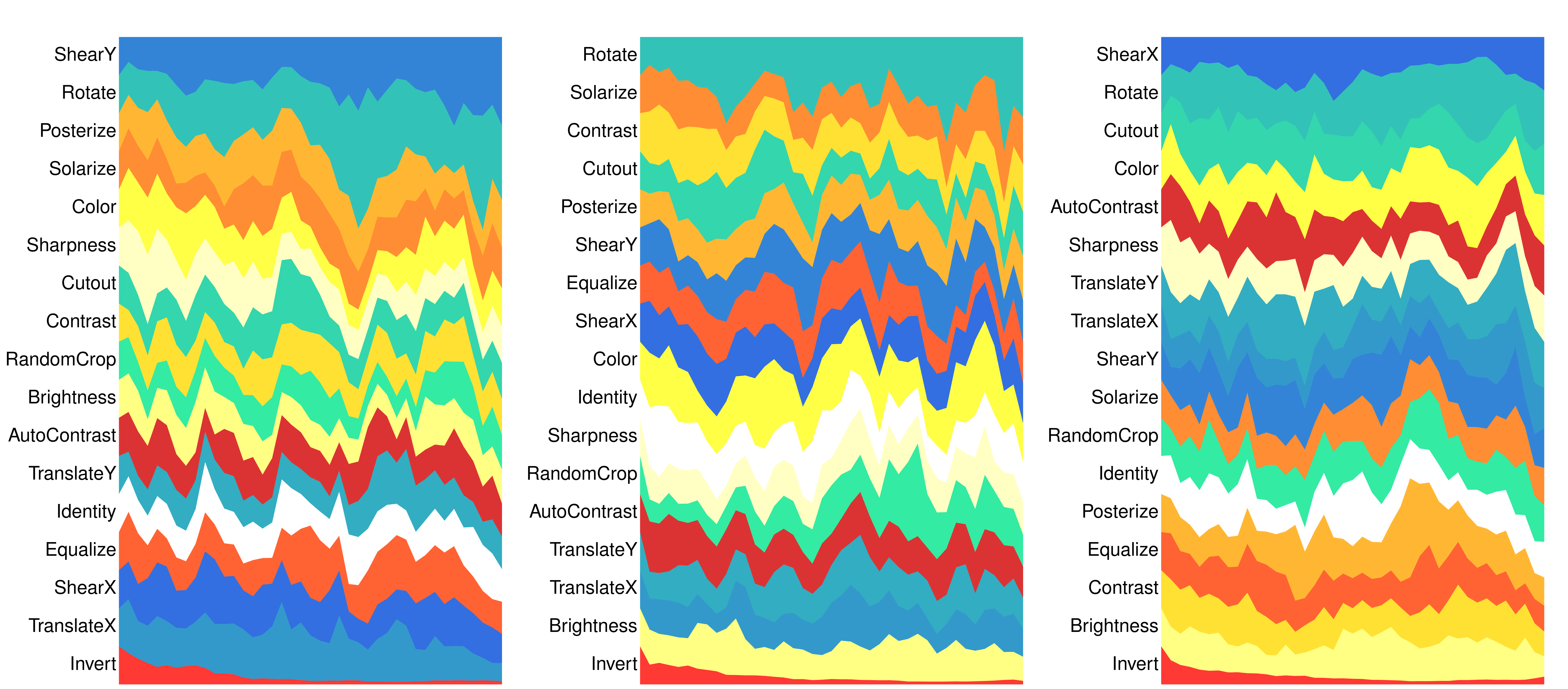}
    \subcaption{The three distributions over transformations forming the composition}      
    \end{subfigure}
    \begin{subfigure}{0.27\linewidth}      \includegraphics[width=\linewidth]{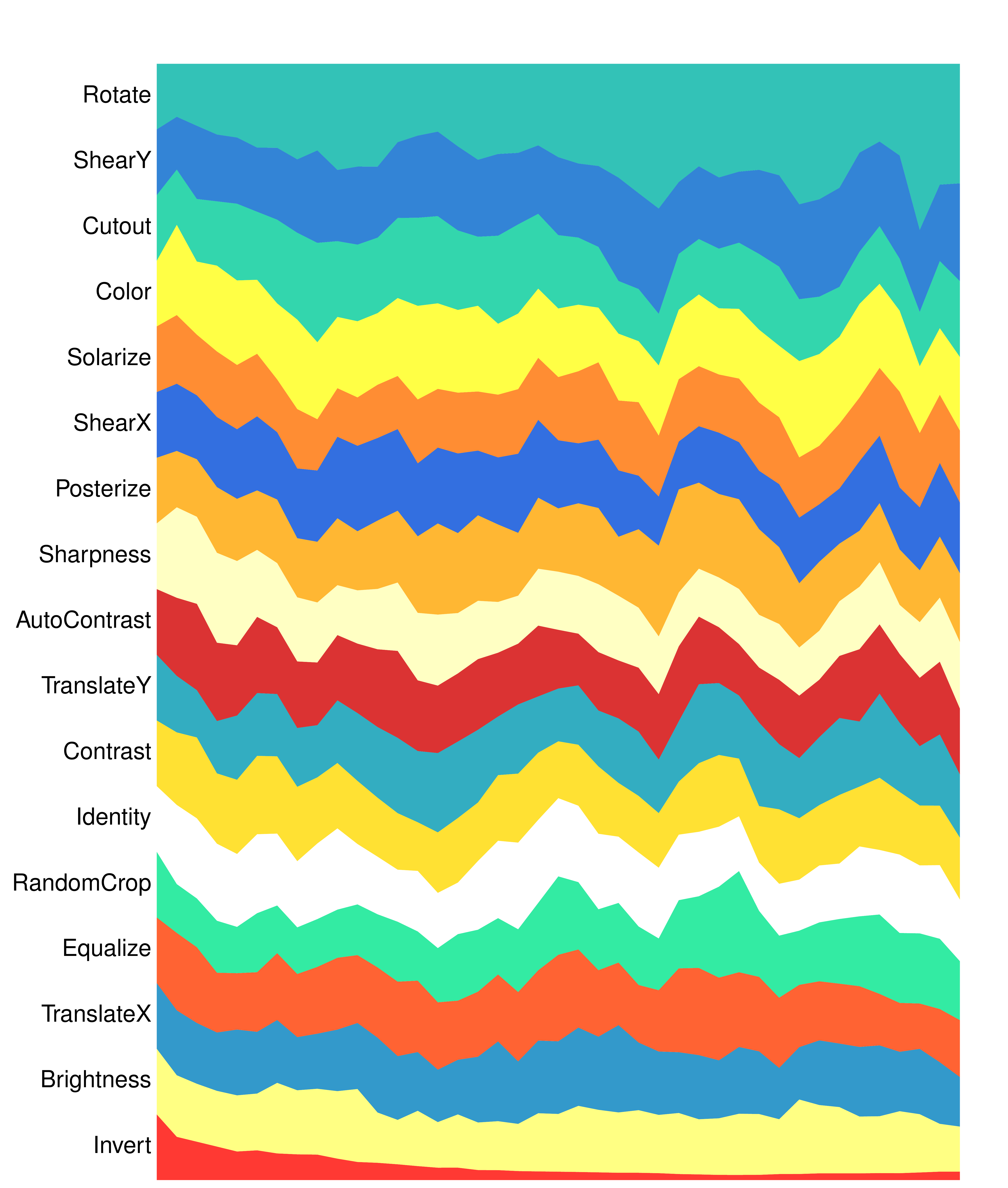}
    \subcaption{Average of the 3 distributions}      
    \end{subfigure}
    \caption{Evolution of the distributions $\pi$ for CIFAR100 with entropy-regularized unrolled optimization, on one of the search splits.}
    \label{fig:ev-unrolled}
\end{figure*}

\begin{figure*}[t!]
    \centering  
    \begin{subfigure}{0.72\linewidth}      \includegraphics[width=\linewidth]{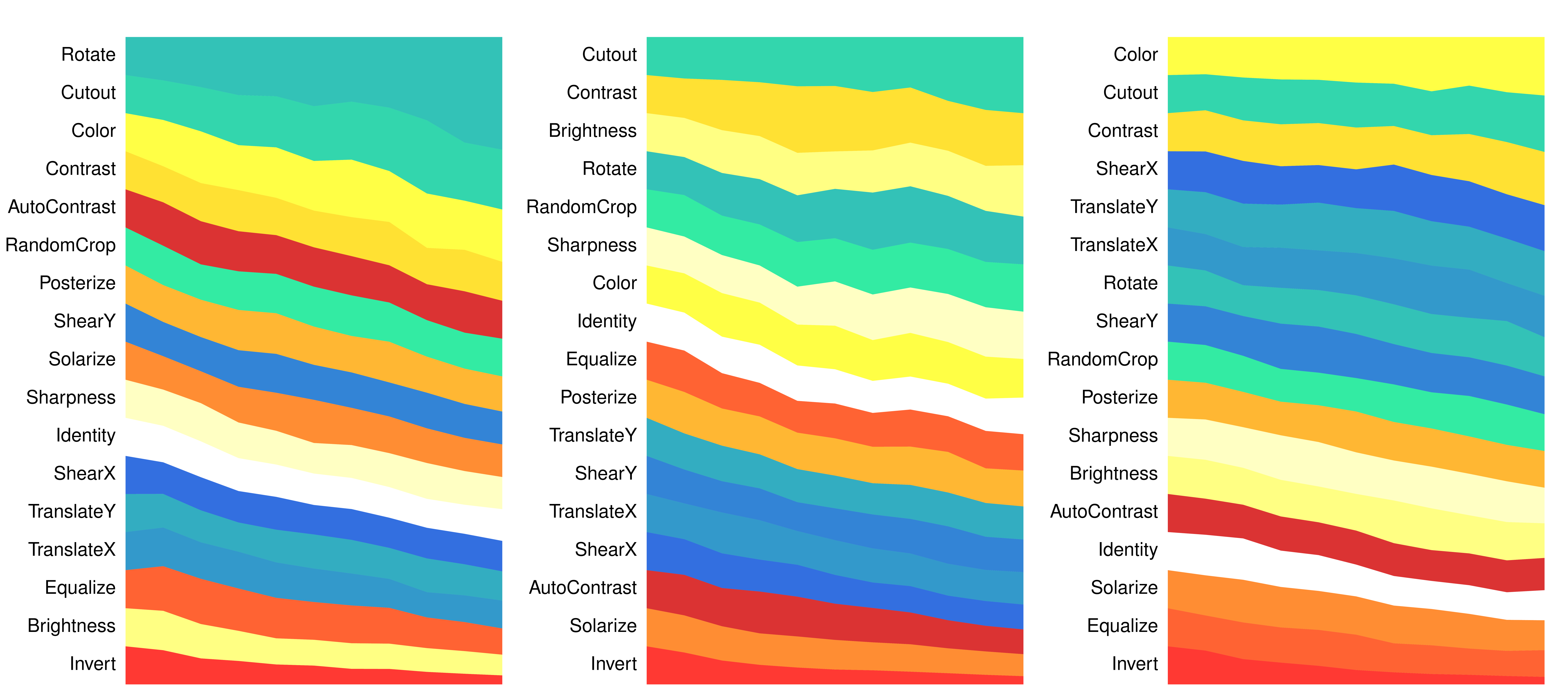}
    \subcaption{0.25: The three distributions over transformations forming the composition}      
    \end{subfigure}
    \begin{subfigure}{0.27\linewidth}      \includegraphics[width=\linewidth]{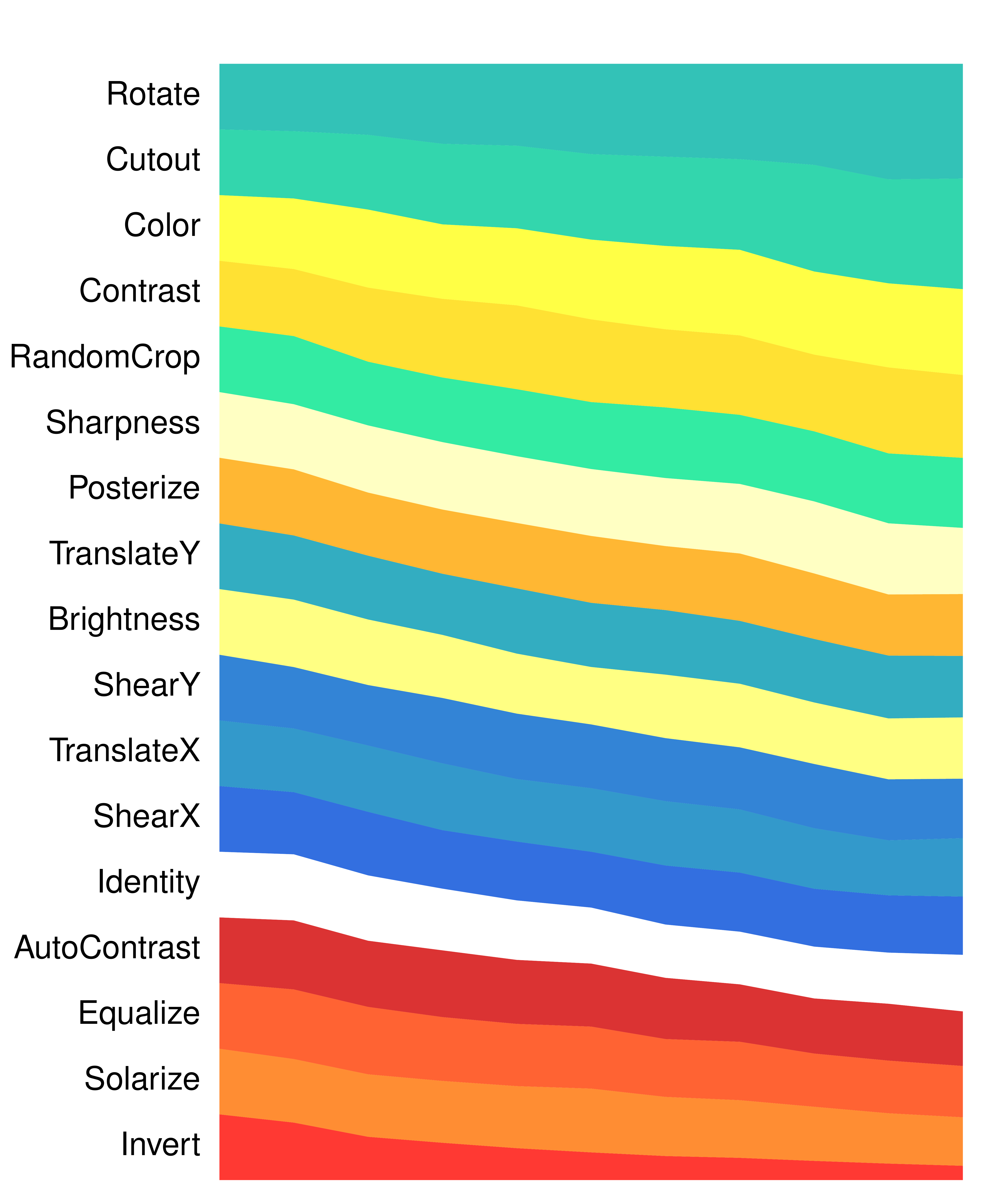}
    \subcaption{0.25: Average of the 3 distributions}
    \end{subfigure} 
    \begin{subfigure}{0.72\linewidth}      
    \includegraphics[width=\linewidth]{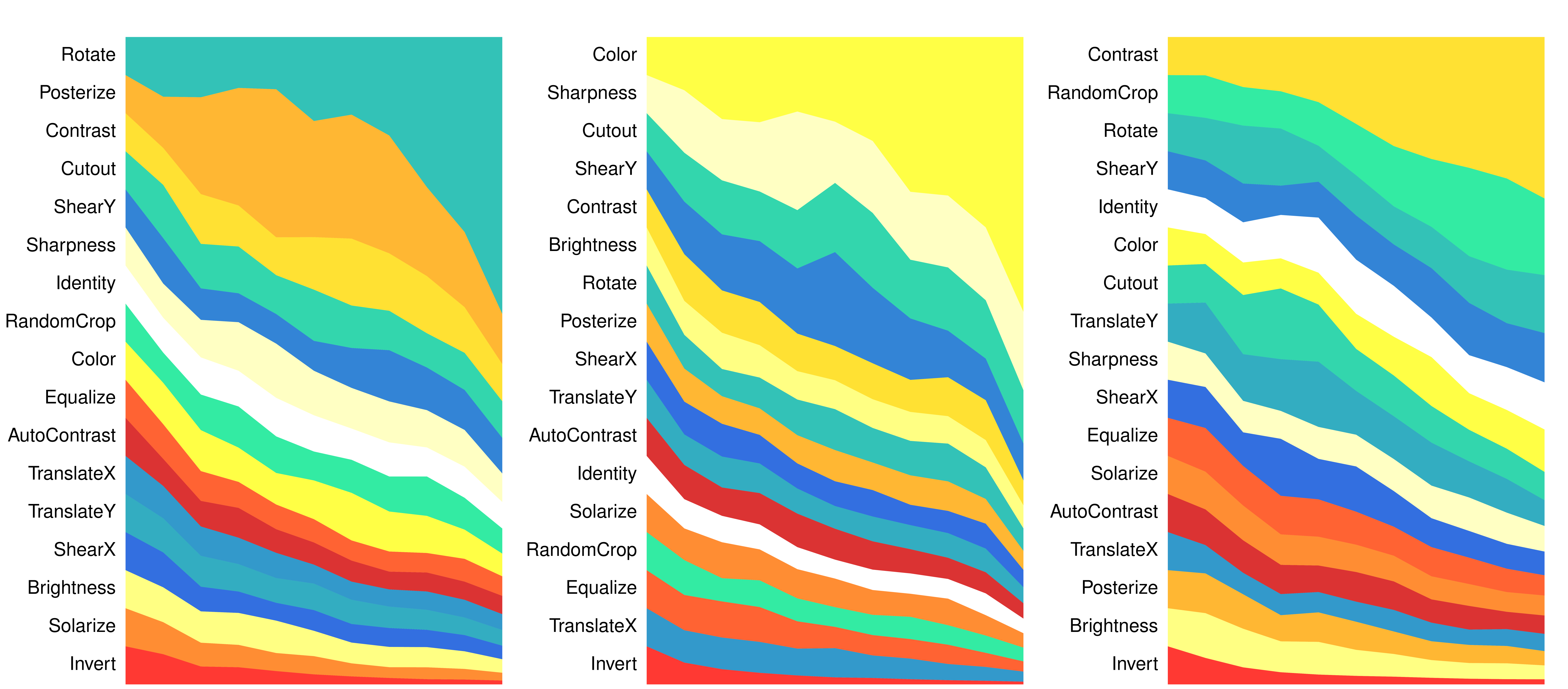}
    \subcaption{0.5: The three distributions over transformations forming the composition}      
    \end{subfigure}
    \begin{subfigure}{0.27\linewidth}      \includegraphics[width=\linewidth]{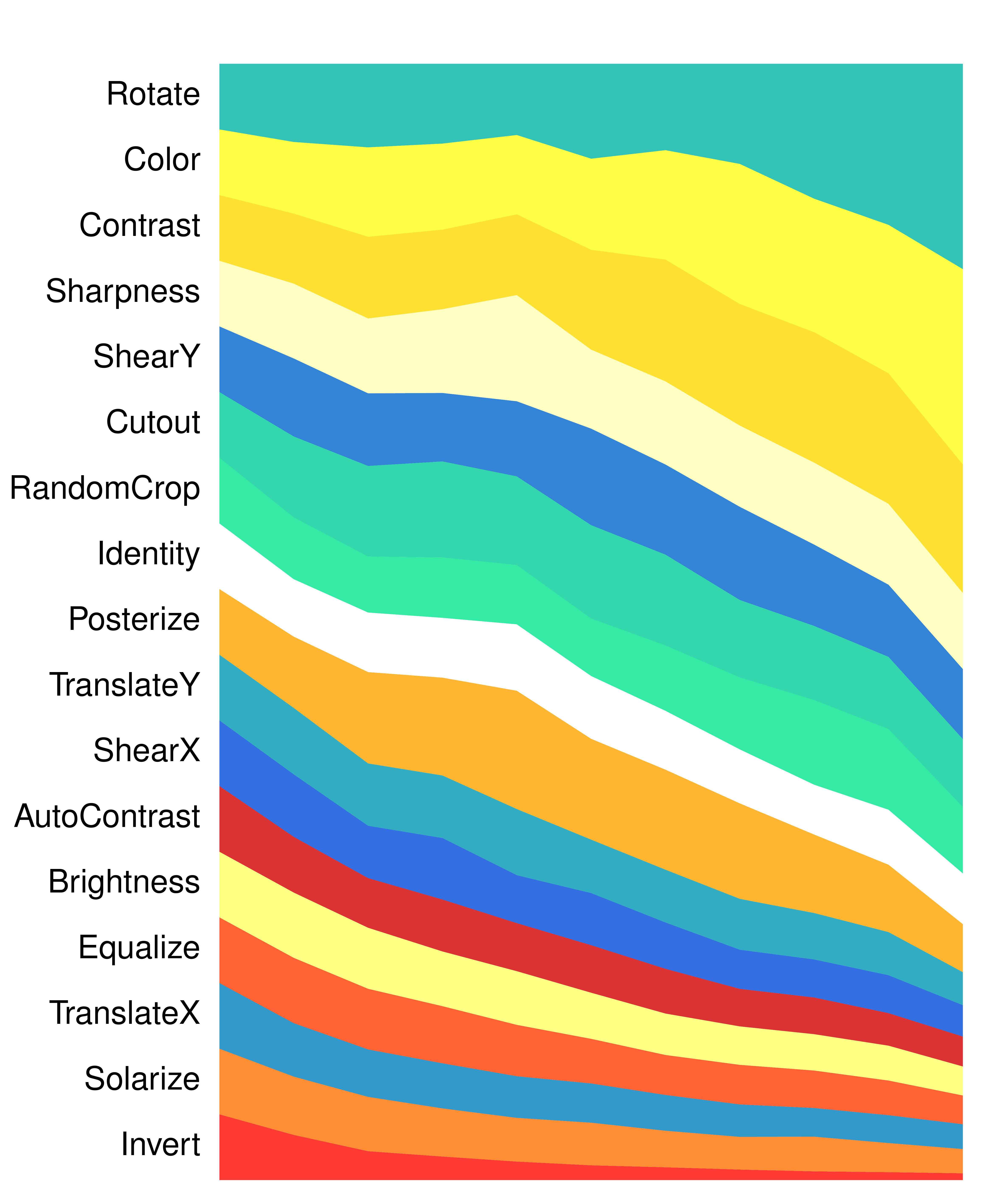}
    \subcaption{0.5: Average of the 3 distributions}      
    \end{subfigure}
    \caption{Evolution of the probability distributions $\pi$ for CIFAR100 with unregularized multi-stage optimization using upper-level learning rates of 0.25 and 0.5.}
    \label{fig:ev-nkl}
\end{figure*}

\myparagraph{Multi-stage optimization without KL regularization}
In our multi-stage approach, $\theta^\star(\tilde\phi)$ is well-approximated at the beginning of each stage, as the model is re-trained with the current policy $\tilde\phi$. Gradient updates close to this policy are `trusted' since our current $\theta$ after re-training stays close to $\theta^\star(\tilde{\phi})$, meaning that we strongly mitigate the approximation inherent to unrolled optimization. Our KL regularization encourages the policy to stay in this \emph{trust region} and without it, the stochasticity of the optimization combined with the high variance from REINFORCE may drive the policy away.
Fig. \ref{fig:ev-nkl} shows the evolution of our probability distributions under an unregularized multi-stage search with two different learning rates, one twice learger than the other. This evolution is smoother than with single-stage unrolled optimization and also quite stable when using a small learning rate, but this slows down convergence, yielding a sub-optimal policy (see Table \ref{tab:ablations}). The larger one leads again to a progressive divergence: the policy is driven too far and the $\theta^\star(\tilde\phi)$ obtained after re-training becomes sub-optimal for the current $\phi$.
In other words, the KL regularization allows making large updates in the parameter space, while remaining close to a reference/anchor policy.

\subsection{An ensembling approach}

In this section, we investigate the effect of an ensembling strategy for SLACK to reduce the variance of gradient updates in the search phase.
More precisely, the strategy consists in independently training multiple models on the lower-level loss while averaging their contributions to the upper-level gradient. Each model is initialized (and subsequently re-initialized at each stage) based on a pre-training with a different seed.
This ensembling strategy was implemented using multiple GPUs, where each GPU trains one copy of the model and only the upper-levels gradients are communicated and averaged across GPUs.

Results on CIFAR10/100 are reported in Table \ref{tab:ablation-ens}.
While  there is a small improvement in most cases, the method still has a strong computational overhead.
Yet it might be a relevant line of research for datasets for which the training procedure has a higher variance, \eg smaller datasets where the additional cost of ensembling is not a significant overhead.

\begin{table*}[t!]
  \centering
  \footnotesize
  \begin{tabular}{lcccccc}
     \toprule
    & \multicolumn{2}{c}{CIFAR10} & \multicolumn{2}{c}{CIFAR100} \\
   \cmidrule(lr){2-3} \cmidrule(lr){4-5} 
      & WRN-40-2 & WRN-28-10 & WRN-40-2 & WRN-28-10 \\ 
    \midrule
    \methodname & 96.29 $\pm$ .08 & 97.46 $\pm$ .06 & 79.87 $\pm$ .11 & 84.08 $\pm$ .16 \\
    Ensembling of  \methodname (4 GPUs) & 96.33 $\pm$ .08 & 97.48 $\pm$ .06 & 79.94 $\pm$ .13 & 84.01 $\pm$ .14 \\
     \bottomrule
\end{tabular}
\caption{CIFAR10/100 accuracy with ensembling strategy.}
\label{tab:ablation-ens}
\end{table*}

\subsection{Warm-start vs cold-start}

In this section, we study the model behaviour when searching with warm-start instead of cold-start. By warm-start, we mean that re-training is performed starting from the current network's weights at the beginning of each stage instead of re-initializing it to its pre-trained weights. 
We experimentally observe that warm-start with the same hyperparameters as for cold-start leads to a progressive overfitting of the network. Increasing the lower-level learning rate mitigates this phenomenon, but still yields sub-optimal results as reported in Table \ref{tab:ablation-warm}. This suggests that re-training from $\theta_0^\star(\phi_{0})$  gives a better estimate of $\theta^\star(\phi_i)$ at stage $i$ than re-training from the biased state close to $\theta^\star(\phi_{i-1})$.

\begin{table*}[t!]
  \centering
  \footnotesize
  \begin{tabular}{lcccccc}
     \toprule
    & \multicolumn{2}{c}{CIFAR10} & \multicolumn{2}{c}{CIFAR100} \\
   \cmidrule(lr){2-3} \cmidrule(lr){4-5} 
      \methodname variant & WRN-40-2 & WRN-28-10 & WRN-40-2 & WRN-28-10  \\ 
      \midrule
    Warm-start & 96.27 $\pm$ .09 & 97.05 $\pm$ .15 & 79.70 $\pm$ .11 & 83.90 $\pm$ .10   \\
    Cold-start (ours) & 96.29 $\pm$ .08 & 97.46 $\pm$ .06 & 79.87 $\pm$ .11 & 84.08 $\pm$ .16 \\
     \bottomrule
\end{tabular}
\caption{CIFAR10/100 accuracy with cold start and warm start.}
\label{tab:ablation-warm}
\end{table*}

\end{document}